\documentclass[10pt,twocolumn,letterpaper]{article}

\usepackage[pagenumbers]{iccv} % To force page numbers, e.g. for an arXiv version

\definecolor{iccvblue}{rgb}{0.21,0.49,0.74}
\usepackage[pagebackref,breaklinks,colorlinks,allcolors=iccvblue]{hyperref}

\usepackage{makecell}
\usepackage[utf8]{inputenc} % allow utf-8 input
\usepackage[T1]{fontenc}    % use 8-bit T1 fonts
\usepackage{url}            % simple URL typesetting
\usepackage{booktabs}       % professional-quality tables
\usepackage{amsfonts}       % blackboard math symbols
\usepackage{nicefrac}       % compact symbols for 1/2, etc.
\usepackage{microtype}      % microtypography
\usepackage{graphicx}
\usepackage{amsmath}
\usepackage{amssymb}
\usepackage{algpseudocode}
\usepackage{algorithm}
\usepackage{multirow}
\usepackage{arydshln}
\usepackage{pifont}
\usepackage{array}
\usepackage{subcaption}
\usepackage[font=small]{caption}
\usepackage{wrapfig}

\usepackage{enumitem}
\usepackage{setspace}

\usepackage{colortbl}
\definecolor{lightblue}{HTML}{EEF4FE}
\definecolor{darkgreen}{rgb}{0.0, 0.5, 0.0}
\usepackage{marvosym}

\def\paperID{7196} % *** Enter the Paper ID here
\def\confName{ICCV}
\def\confYear{2025}

\title{Flash-VStream: Efficient Real-Time Understanding for Long Video Streams}

\author{
Haoji Zhang$^{1,2}$\thanks{Equal contribution.\ \ \ \ \ \ 
\textsuperscript{\Letter}Corresponding authors.\ \ \ \ \ \ 
$\dagger$Project leader.
}\quad
Yiqin Wang$^{1,2*}$\quad
Yansong Tang$^{1,2}$\textsuperscript{\Letter}\quad
Yong Liu$^{1,2}$\quad
Jiashi Feng$^{4}$\quad
Xiaojie Jin$^{3,4}$\textsuperscript{\Letter}$^{\dagger}$
\\
{\normalsize 
$^{1}$Tsinghua University\quad
$^{2}$Tsinghua Shenzhen International Graduate School\quad
$^{3}$Beijing Jiaotong University \quad
$^{4}$ByteDance Inc.
}
}

\begin{document}
    \maketitle
    
\begin{abstract}
Benefiting from the advances in large language models and cross-modal alignment, existing multimodal large language models have achieved prominent performance in image and short video understanding. 
However, the understanding of long videos is still challenging, as their long-context nature results in significant computational and memory overhead. Most existing work treats long videos in the same way as short videos, which is inefficient for real-world applications and hard to generalize to even longer videos.
To address these issues, we propose Flash-VStream, an efficient video language model capable of processing extremely long videos and responding to user queries in real time. 
Particularly, we design a Flash Memory module, containing a low-capacity context memory to aggregate long-context temporal information and model the distribution of information density, and a high-capacity augmentation memory to retrieve detailed spatial information based on this distribution.
Compared to existing models, Flash-VStream achieves significant reductions in inference latency.
Extensive experiments on long video benchmarks and comprehensive video benchmarks, i.e., EgoSchema, MLVU, LVBench, MVBench and Video-MME, demonstrate the state-of-the-art performance and outstanding efficiency of our method.
Code is available at \url{https://github.com/IVGSZ/Flash-VStream}.
\end{abstract}

\vspace{-10pt}    
    \section{Introduction}
\label{sec:intro}
% 纵览全局，long video的重要性
%In the pursuit of Artificial General Intelligence (AGI), the ability to perceive and understand complex, dynamic environments stands as a pivotal milestone.
%An important target of this journey is to develop advanced large multimodal 
%models~\cite{achiam2023gpt4report,team2024gemini,wang2024qwen2vl}, which are designed to function as powerful and responsive AI assistants, handling diverse sources such as text, images, audio, and video. 
% Achieving robust perception and understanding of complex, dynamic environments is a critical milestone toward Artificial General Intelligence (AGI). A key objective in this direction is developing advanced large multimodal models~\cite{achiam2023gpt4report,team2024gemini,wang2024qwen2vl} that effectively handle various types of data, such as text, vision and audio.
Achieving robust perception and understanding of complex, dynamic environments is a crucial milestone toward Artificial General Intelligence (AGI). A core objective in this pursuit is the development of advanced large multimodal models~\cite{achiam2023gpt4report,team2024gemini,wang2024qwen2vl} capable of effectively integrating diverse data types, including text, visual content, and audio.

\begin{figure}[t]
    \centering
    \includegraphics[width=1\linewidth]{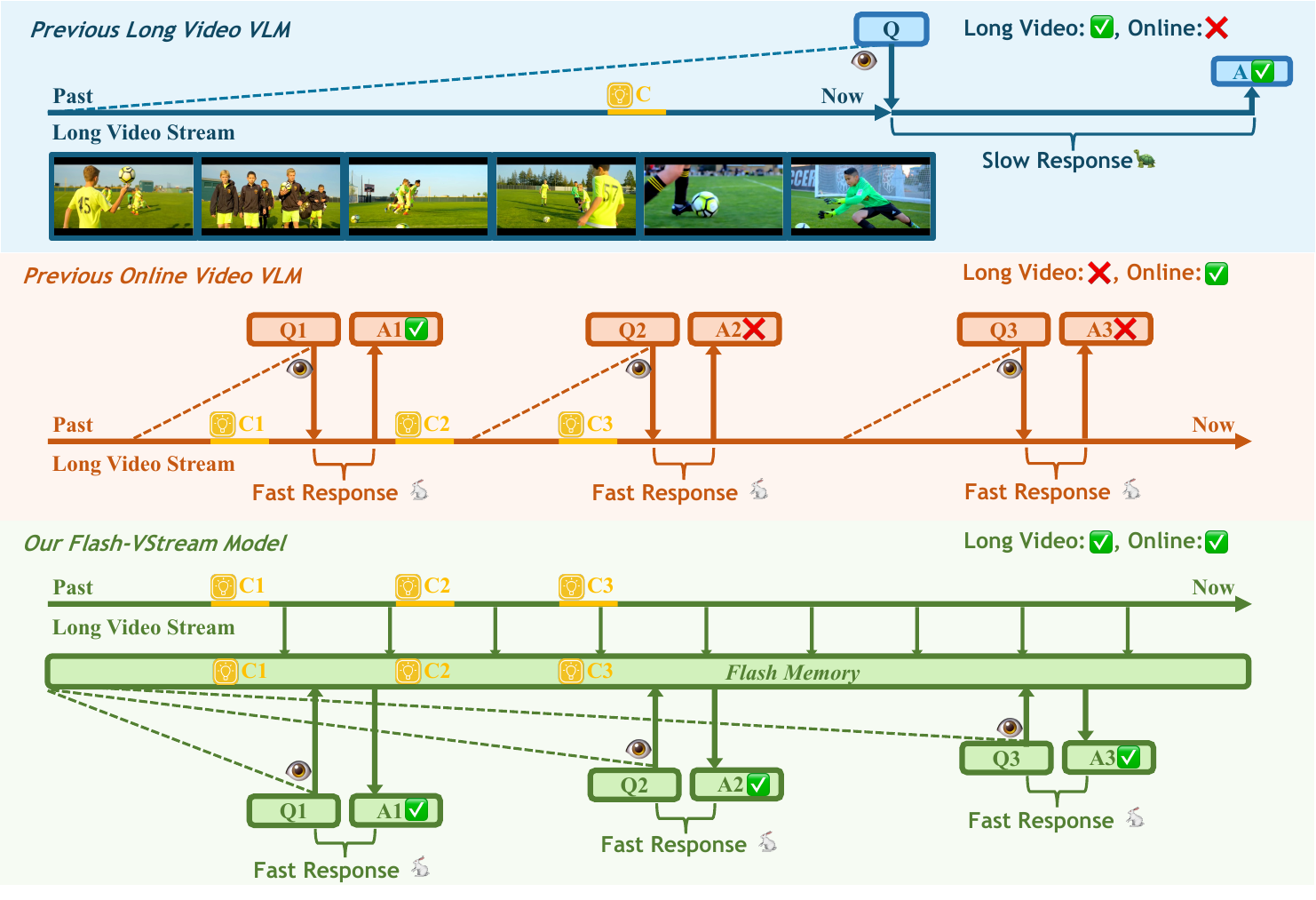}
    \vspace{-20pt}
    \caption{
        \textbf{Comparison with previous methods.}
        Flash-VStream can understand long videos accurately in an online manner.
        Here ``C'' denotes critical clues for questions.
    }
    \vspace{-10pt}
    \label{fig:teaser} 
\end{figure}

% 为什么long video理解的efficiency很重要
Among multimodal tasks, long video understanding stands out as particularly significant yet challenging, primarily due to its substantial computational overhead and high GPU memory demands. Consequently, improving model efficiency is essential for making long video understanding practical. High GPU memory usage limits real-world deployment, particularly on resource-constrained edge devices. Additionally, excessive computational requirements increase inference latency, directly affecting applications requiring real-time human-computer interactions. In this context, we define a video language model (VLM) as \textit{real-time} if it can respond to user queries within one second.
% Among these, understanding long videos is particularly important yet challenging, due to its huge memory and computational overhead. Therefore, model efficiency is crucial for long video understanding tasks.
% The costs of GPU memory and computation are two critical factors.
% %that must be considered, as highlighted by~\cite{moviechat}.
% Specifically, GPU memory costs impede real-world deployments, especially on edge devices.
% The computation cost influences inference latency, which is a requirement of real-time human-computer interaction. 
% In this context, we define a video language model (VLM) as \textit{real-time} if it can respond to user queries within one second.

% 为什么long video理解的efficiency很重要：real-time HCI的广阔应用
% Real-time human-computer interaction is necessary in many scenarios. 
% As mentioned above, multimodal assistants~\cite{achiam2023gpt4report,team2024gemini,wang2024qwen2vl} are expected to have real-time capabilities for more fluent conversations.
% In robotics, robots operating in the wild can leverage VLMs to interpret and react to their environment in real-time~\cite{supancic2017tracking, sermanet2023robovqa}.
% Similarly, in surveillance systems, real-time VLMs can process and analyze video streams from specific locations continuously, ensuring overall security~\cite{chen2019distributed, muhammad2019deepres}. 
% % todo: 在网络直播领域
% However, currently even the best VLMs still fails to perform real-time long video question-answering upon user queries~\cite{vistallama,chen2024sharegpt4video,li2024llavaov,wang2024qwen2vl}.
Real-time interaction capabilities are vital in numerous practical scenarios. Multimodal assistants~\cite{achiam2023gpt4report,team2024gemini,wang2024qwen2vl,wang2024ponder}, for example, must operate in real-time to ensure fluid user experiences. 
Similarly, robots deployed in real-world environments benefit from VLMs capable of interpreting and reacting swiftly to dynamic situations~\cite{supancic2017tracking,sermanet2023robovqa}. 
Surveillance systems also rely on real-time VLMs to continuously analyze video streams, thereby maintaining security effectively~\cite{chen2019distributed,muhammad2019deepres,gan2023temporal}. 
However, current state-of-the-art VLMs still struggle to achieve real-time responsiveness when performing question-answering tasks on long videos~\cite{vistallama,chen2024sharegpt4video,li2024llavaov,wang2024qwen2vl}.

% 我们的idea
% To address this challenge, we starts with a widely corroborated observation, that \textit{temporal redundancy} exists in all forms of 
% videos~\cite{wang2021adaptive,wang2022adafocus,wang2022adafocusv3,
% ghodrati2021frameexit,korbar2019scsampler,sun2021dynamic,han2021dynamicsurvey}.
% Since not all frames are equally important for video understanding, models are expected to allocate more computation to more informative frames and avoid processing very similar frames.
% In~\Cref{sec:flash}, we propose the Flash Memory module.
To address this challenge, we start from a widely recognized observation: \textit{temporal redundancy} is prevalent in all video types~\cite{wang2021adaptive,wang2022adafocus,wang2022adafocusv3,ghodrati2021frameexit,han2021dynamicsurvey}. Since not all video frames are equally informative, an efficient model should allocate more computational resources preferentially to critical frames. In \Cref{sec:flash}, we propose the Flash Memory module, which addresses these challenges by integrating a Context Synopsis Memory that captures information density distribution along the temporal dimension, and a Detail Augmentation Memory designed to retrieve detailed content by selectively sampling key frames.

% As illustrated in~\Cref{fig:flash_memory}, Flash-VStream utilizes a two-process asynchronous framework to separate vision processing from language processing. 
In this paper, we introduce Flash-VStream, a two-process video-language model capable of efficiently handling extremely long video streams and providing real-time responses to user queries.
As illustrated in~\Cref{fig:flash_memory}, the frame handler operates continuously, encoding new frames through a visual encoder and updating the Flash Memory without interruption. Concurrently, the question handler functions as a server process, triggered by incoming questions, and leverages Flash Memory to rapidly generate the first token of the answer within one second.
In summary, the two-process framework ensures simultaneous video processing, memory updating and real-time response generation.
The innovative memory design sets our model apart from previous works, in terms of video length and online capability (~\Cref{fig:teaser}).

\begin{figure}[t]
    \centering
    % 量化的图
    \includegraphics[width=1\linewidth]{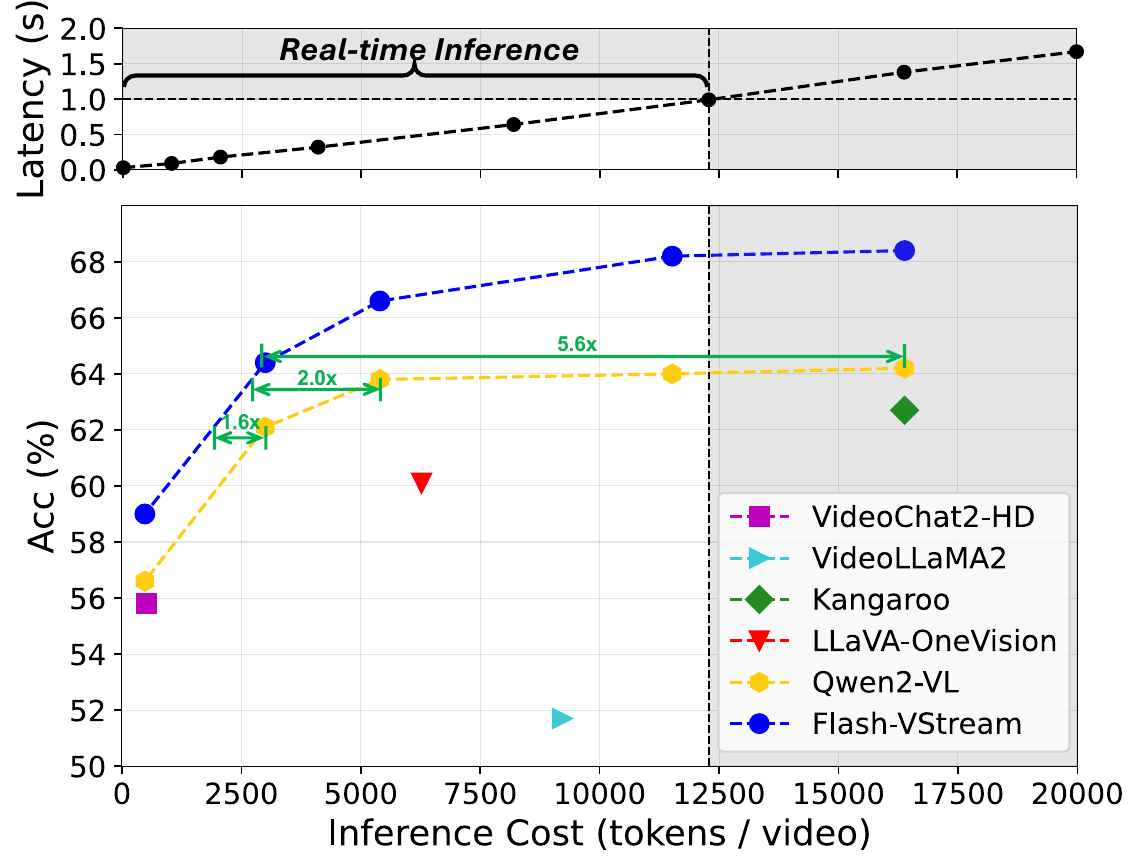}
    \vspace{-10pt}
    \caption{
        \textbf{Response latency / Accuracy on EgoSchema v.s. Inference cost.}
        % Latency is tested on an A100 GPU.
        Inputting more than 12000 tokens will not meet real-time requirements.
        Flash-VStream can respond to user queries in real time while maintaining outstanding performance.
        % Flash-VStream achieves 
        }
    \vspace{-10pt}
    \label{fig:efficiency}
\end{figure}

\begin{figure*}[t]
    \centering
    \includegraphics[width=1\linewidth]{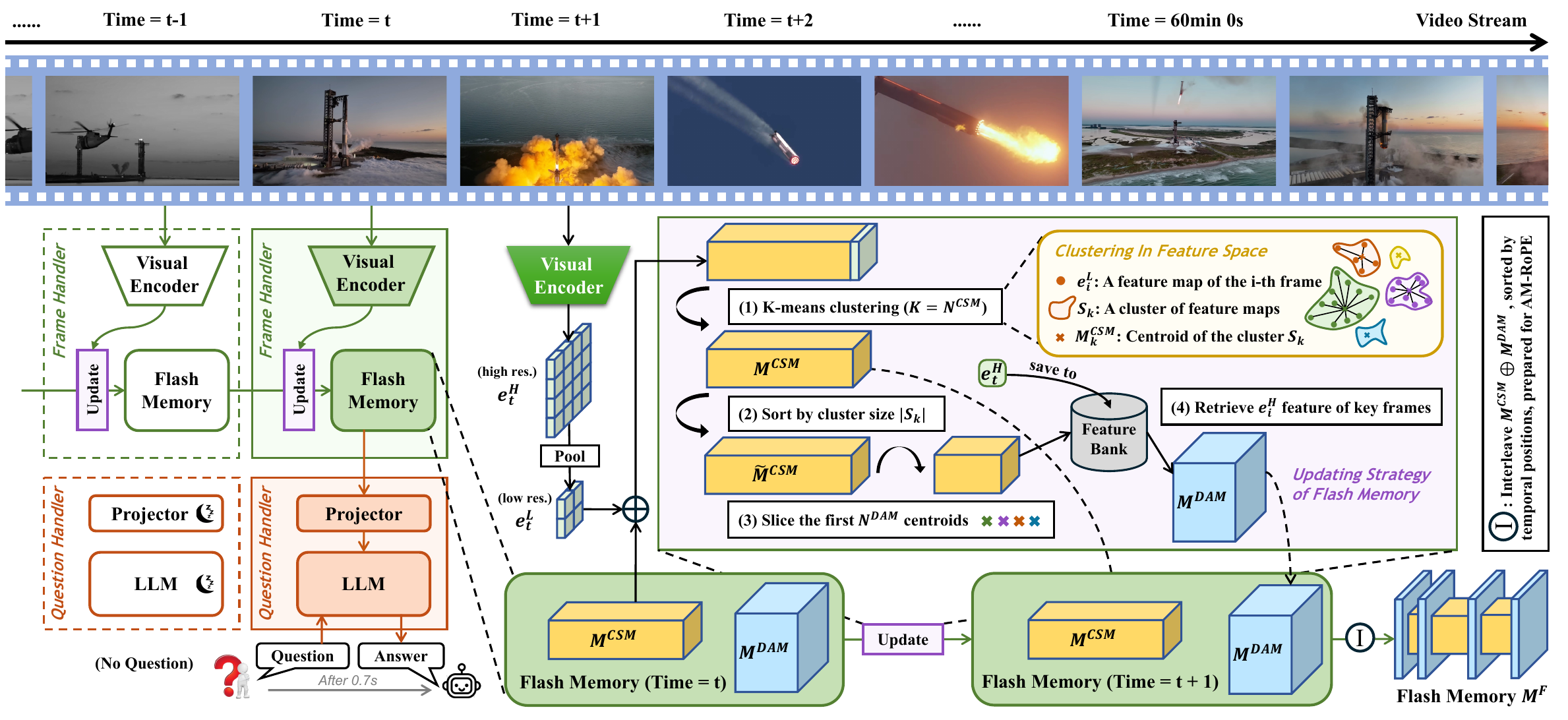}
    \caption{\textbf{Overview of Flash-VStream two-process framework.}
        The frame handler process continuously encodes new frames.
        The question handler process asynchronously responds to human inquiries in real-time.
        Flash Memory is composed of interleaved Context Synopsis Memory and Detail Augmentation Memory,
        organized in chronological order. 
        CSM is updated by clustering low resolution feature maps on an inter-frame level.  
        DAM is updated by retrieving high resolution feature maps of the most informative frames from a feature bank. 
    }
    \label{fig:flash_memory} 
    \vspace{-10pt}
\end{figure*}

As demonstrated in~\Cref{fig:efficiency}, Flash-VStream significantly reduces inference latency to meet real-time standards, achieving a superior balance between accuracy and efficiency, and setting new state-of-the-art performance on the full EgoSchema benchmark~\cite{egoschema}.

We further validate the generalization capability of Flash-VStream through zero-shot video question answering experiments on three long video understanding benchmarks (EgoSchema~\cite{egoschema}, MLVU~\cite{zhou2024mlvu}, LVBench~\cite{wang2024lvbench}) and two comprehensive video understanding benchmarks (MVBench~\cite{li2024mvbench_videochat2}, Video-MME~\cite{fu2024videomme}), as shown in~\Cref{tab:maintab}. Additionally, detailed ablation studies in~\Cref{sec:exp_ablation} clearly verify the effectiveness of Flash Memory.
We summarize our contributions as follows:
\begin{itemize}[leftmargin=*]
\item We introduce Flash-VStream, an efficient large video language model capable of processing extremely long video and providing real-time responses to user queries. 
% Flash-VStream employs a two-process asynchronous framework to decouple vision processing and language processing, ensuring efficient and timely responses.
Flash-VStream utilizes a fixed-size Flash Memory to bridge a two-process framework, ensuring efficient and timely processing of long-term video streams.

\item A novel Flash Memory module is proposed to reduce temporal redundancy between consecutive frames, which includes a Context Synopsis Memory to model the distribution of information density along the temporal dimension and a Detail Augmentation Memory to retrieve detailed information from key frames.

\item Extensive experiments and ablation studies on long video understanding benchmarks and comprehensive video understanding benchmarks demonstrate the outstanding performance and efficiency of Flash-VStream.
\end{itemize}

\section{Related Work}
\label{sec:related}

% main method

% 讲一下MLLM must
\subsection{Video Language Models}
With recent advances in Large Language Models (LLMs)~\cite{brown2020language,Ouyang2022TrainingLM,Touvron2023Llama2O,touvron2023llama,dubey2024llama3,yang2024qwen2}, and Multimodal Large Language Models 
(MLLMs)~\cite{Li2022BLIPBL,li2023blip,Dai2023InstructBLIPTG,llava,Liu2023ImprovedBW,li2024llavaov,team2025kimivl},
many works begin to build Video Language Models (VLMs) based on them.
LLaMA-VID~\cite{llamavid} represents single-frame features with only 2 tokens.
Chat-UniVi~\cite{chatunivi} employs dynamic tokens to model video features of different scales.
Other works~\cite{vistallama,ye2025voco,ye2025atp,wang2024hierar} use different compression techniques to represent an entire video with fewer tokens.
While these methods succeeded in short video understanding, they have relatively poor performance on long video understanding benchmarks~\cite{fu2024videomme,egoschema}.

% 对比其他long video understanding method must
\subsection{Long Video Understanding}
Earlier work MIST~\cite{gao2023mist} introduces an iterative method to select the most question-related video clip.
SEVILA~\cite{yu2024sevila} performs temporal key frame localization and video question answering simultaneously.
Compared to question-aware methods~\cite{llamavid,gao2023mist,yu2024sevila}, Flash-VStream does not rely on specific questions and can achieve general information aggregation based on visual information itself. 
MovieChat~\cite{moviechat} proposes to merge similar frame features by average pooling.
Though it is able to process long video with limited GPU memory cost, its performance is suboptimal due to its training-free framework.
Recent works~\cite{zhang2024longva,xue2024longvila} explore long context extension finetuning. However, they are computationally expensive.
RETAKE~\cite{wang2024retake} proposes a KV cache pruning method, which successfully lowers the knowledge redundancy in long videos.
% RETAKE~\cite{wang2024retake} designs a KV cache pruning method and success to lower knowledge redundancy in long videos. 
% LongVA~\cite{zhang2024longva} claims that finetuning on long text data and short video data could enhance long video understanding ability of models.
% LongVILA~\cite{xue2024longvila} designs a Multi-Modal Sequence Parallelism algorithm for efficient long video training and inference. 
Different from them, Flash-VStream keeps most informative frames in memory to reduce temporal redundancy, resulting in higher efficiency.

% 对比其他的streaming method must
\subsection{Real-Time Video Stream Understanding}
Real-time video stream understanding requires models to process video streams and finish specific tasks in real-time.
Most existing methods are designed to perform a specific vision task, such as real-time object tracking~\cite{track1,track3},
action recognition~\cite{action1,action2,wang2024uniada},
and segmentation~\cite{luo2023soc,yang2024lavt,wang2025iterprime}.
Zhou et al.~\cite{zhou2024streamingcaption} design a streaming framework for dense video captioning.
VideoLLM-Online~\cite{chen2024videollmonline} is designed for streaming video narration and action anticipation.
Considering natural language is becoming a general interface for various modalities~\cite{gao2021clip,li2023blip,wang2025sam2love} and tasks~\cite{multimodal3,bai2024self,zhu2025instarevive},
our work focuses on real-time video question answering (VQA) upon free-form user queries, which is more challenging.
% a more challenging and general task~\cite{yu2019activitynet,xiao2021next}.

% 对比其他领域的memory方法
\subsection{Memory for Long Sequence Modeling}
Memory mechanisms are widely used to store and retrieve information in all forms of long sequence modeling tasks,
such as time series forecasting~\cite{chang2018memory}, recommendation system~\cite{tan2021dynamic}, 
% machine translation~\cite{Daelemans_2005}, 
and video object segmentation~\cite{cheng2022xmem,track2}.
For video understanding, MovieChat~\cite{huang2020movienet} utilizes a long-term memory and short-term memory framework. 
MC-ViT~\cite{balavzevic2024mcvit} proposes a memory-consolidated vision transformer for long video understanding.
GLMGIR~\cite{yan2019finegrain} proposes a multi-granularity memory to solve fine-grained video captioning.
In comparison, our method uses two memory modules that focus on temporal information modeling and spatial detail enhancement, resulting in a synergic improvement of efficiency.

    \section{Flash-VStream}
\label{sec:flash}

Flash-VStream improves model efficiency by allocating more computation to the most informative and representative key frames.
As illustrated in~\Cref{fig:flash_memory}, Flash Memory is updated iteratively to retain key information from both current and historical frames. 
The Flash Memory comprises a Context Synopsis Memory (CSM) for long-term temporal information aggregation and representative key frame localization. 
It also contains a Detail Augmentation Memory (DAM) for enhancing the spatial details of the key frames.

% overall arch, 2-process
\subsection{Model Architecture}
In order to lower the inference latency, Flash-VStream decouples vision processing and language processing into two processes.
As presented in~\Cref{fig:flash_memory}, the frame handler process is responsible for continuously frame encoding and memory consolidation,
while the question handler process remains online, waiting for user queries.
These processes collaborate by reading from and writing to shared memory, namely the Flash Memory.
Following common practices, we adopt a Vision Transformer~\cite{dosovitskiy2020image} as the visual encoder, a 2-layer MLP~\cite{Liu2023ImprovedBW} as the projector, and a Qwen2-7b LLM~\cite{yang2024qwen2} as the language decoder.
Formally, the visual encoder takes the $t$-th frame 
$V_t \in \mathbb{R}^{H\times W\times 3}$ as input and outputs a high resolution feature map 
$e_t^{\text{H}} \in \mathbb{R}^{h\times w\times d}$ and a low resolution feature map $e_t^{\text{L}} \in \mathbb{R}^{h^\prime\times w^\prime\times d}$: 
\begin{align}
e_t^{\text{H}} &= f_{\text{enc}}(V_t) \\
e_t^{\text{L}} &= f_{\text{enc}}(\operatorname{pool}(V_t))
\end{align}
Here $h\times w, h^\prime\times w^\prime$ denotes the number of patches, $d$ represents the hidden dimension size of visual encoder and ``pool'' stands for an average pooling layer with ratio $R_\text{pool}=4$.

\begin{table}[h]
    \centering    
    \footnotesize
    \setlength{\tabcolsep}{8pt}
    \begin{tabular}{l|c|cc}
        \Xhline{0.8px}
            \rowcolor{lightblue} \textbf{Configuration} & \textbf{Notation} & \textbf{CSM} & \textbf{DAM} \\
        \hline
            Input Frames  &     -    & $120$    & $60$ \\
            Input Resolution & $H\times W$ & $224\times 224$  & $448\times 448$ \\
            Temporal Size & $N$         & $60$    & $30$ \\
            Spatial Size  & $h\times w$ & $256$    & $1024$ \\
            LLM tokens    & $N_{\text{Vtokens}}$ & $60\times 64$ & $30\times 256$ \\
        \Xhline{0.8px}
    \end{tabular}
    \caption{\textbf{Flash Memory configurations.} 
    The shape of an input frame can be rectangular, as long as pixel amount is less than Input Resolution.
    Temporal Size is the number of feature maps of memory. Spatial Size is the number of ViT tokens in a feature map.
    }
    \label{tab:config}
    \vspace{-10pt}
\end{table}

% CSM
\subsection{Context Synopsis Memory}
\label{sec:flash_csm}
CSM is designed for efficient long-term understanding.
To represent as many frames as possible within limited resources,
CSM ($M^{\text{CSM}}$) uses a set of compressed low resolution feature maps with size $S^{\text{CSM}}=N^{\text{CSM}}\times h^\prime\times w^\prime\times d$. 
% Ideally, all frames in a long video should be processed, so low resolution feature maps are more suitable for better efficiency.
CSM integrates semantically similar frames, which form a cluster of similar frames, i.e., the \textit{context}. 
As defined in \Cref{eq:csm}, each item in CSM is the centroid of a cluster of low resolution feature maps, i.e., the \textit{synopsis} of the context.
\vspace{-5pt}
\begin{align}
M^{\text{CSM}} &= \left\{ \frac{1}{|S_k|} \sum_{i \in S_k} e_i^{\text{L}} \right\}_{k=1}^{N^\text{CSM}}, 1 \leq i \leq t \label{eq:csm} \\
M^{\text{CSM}} &=\operatorname{cluster} ( M^{\text{CSM}} \oplus e_{t+1}^{\text{L}})\label{eq:cluster}
\end{align}
Here $S_k$ is the k-th cluster set with size=$|S_k|$, and $\oplus$ denotes feature concatenation.
CSM is initialized with the feature maps of the first $N^{\text{CSM}}$ frames.
In~\Cref{eq:cluster}, CSM is updated by clustering algorithm using a new feature map of the $(t+1)$-th frame, following~\cite{chatunivi,balavzevic2024mcvit,moviechat}.
Since we need to limit the number of clusters to $N^{\text{CSM}}$ 
, K-means clustering~\cite{macqueen1967some} is employed as an effective and efficient clustering algorithm. 
% \textcolor{blue}{C is easily misrecognized as dimension, can use text CSM. The same for ND in Sec. 3.3.}

Therefore, CSM maintains the cluster centroids to reduce additional computation, and the clusters themselves serve as an implicit representation of information density.

% DAM
\subsection{Detail Augmentation Memory}
\label{sec:dam}
% While CSM collects long-term information, it ignores many spatial details in videos.
% This is where Detail Augmentation Memory (DAM) becomes essential.
While CSM effectively aggregates long-term temporal information, it compromises spatial details critical for fine-grained video understanding.
To address this, we propose DAM as a complementary module to retain and augment spatial details crucial for precise video understanding. 
% Since high resolution feature maps are computationally expensive, DAM operates by selectively storing high resolution feature maps of key frames.
Considering the prohibitive computational costs of high resolution feature maps, DAM operates by selectively storing high resolution feature maps of key frames.

It is natural to borrow the results of CSM for free and use them to guide key frame selection.
Given that a CSM cluster centroid $M^{\text{CSM}}_k$ is a synopsis of multiple frames,
DAM retrieves fine-grained spatial features based on these centroids.
DAM adopts a \textit{Feature-Centric} retrieval policy for key frame localization.
Specifically, DAM is a set of high resolution feature maps 
$M^{\text{DAM}}$ with size $S^{\text{DAM}}=N^{\text{DAM}}\times h\times w\times d$. 
\begin{align}
    \tilde{M}^{\text{CSM}} &= \operatorname{sort} (M^{\text{CSM}}, \operatorname{key=}\{|S_k|\}_{k=1}^{N^\text{CSM}}) \label{eq:sort} \\
    M^{\text{DAM}} &= \{ e_{f(k)}^{\text{H}} \}_{k=1}^{N^\text{DAM}},
    f(k) =\underset{i}{\operatorname{argmin}}\, D(\tilde{M}^{\text{CSM}}_{k}, e_i^{\text{L}}) \label{eq:dist}
\end{align}
As shown in~\Cref{fig:flash_memory}, DAM takes the top-$N^{\text{DAM}}$ largest cluster centroids as anchors. A frame is considered a key frame if its $e_i^{\text{L}}$ is the nearest to a centroid anchor in the feature space. 
In~\Cref{eq:sort,eq:dist}, $\tilde{M}^{\text{CSM}}_{k}$ represents the centroid of the $k$-th largest cluster, sorted by the cluster size $|S_k|$, i.e., the number of frames in the cluster. 
$f(k)$ denotes the index of the $k$-th important key frame. 
$D(\cdot,\cdot)$ denotes the Euclidean distance function.
More ablation studies are provided in~\Cref{sec:exp_ablation}.

A Feature Bank 
$E^{\text{H}}_t=\{e_{1}^{\text{H}}, e_{2}^{\text{H}}, ..., e_{t}^{\text{H}}\}$ 
is maintained for the retrieval of high resolution feature maps, where $t$ is the number of current frames.
The feature bank can be offloaded to disk to avoid memory overflow.

In short, CSM captures long-term temporal information and DAM augments it with more fine-grained spatial details of key frames. They complement each other to provide a comprehensive understanding of long videos.
As shown in~\Cref{fig:flash_memory}, Flash Memory $M^F$ is the interleaved form of CSM and DAM, sorted by temporal positions of feature maps. 
Formally, temporal positions can be calculated as:
\vspace{-5pt}
\begin{align}
    P^{\text{CSM}}_k &= \frac{1}{|S_k|} \sum_{i \in S_k} i, 1 \leq k \leq  N^{\text{CSM}} \\
    P^{\text{DAM}}_k &= f(k), 1 \leq k \leq N^{\text{DAM}} \\
    M^F =\operatorname{sort}(M^{\text{CSM}} &\oplus M^{\text{DAM}}, \operatorname{key}=
    P^{\text{CSM}} \oplus P^{\text{DAM}})
\end{align}

\begin{figure}[t]
    \centering
    % 量化的图
    % \includegraphics[width=1\linewidth]{pic/capacity_5.pdf}
    \begin{subfigure}[b]{0.49\linewidth}
        \centering
        \includegraphics[width=\linewidth]{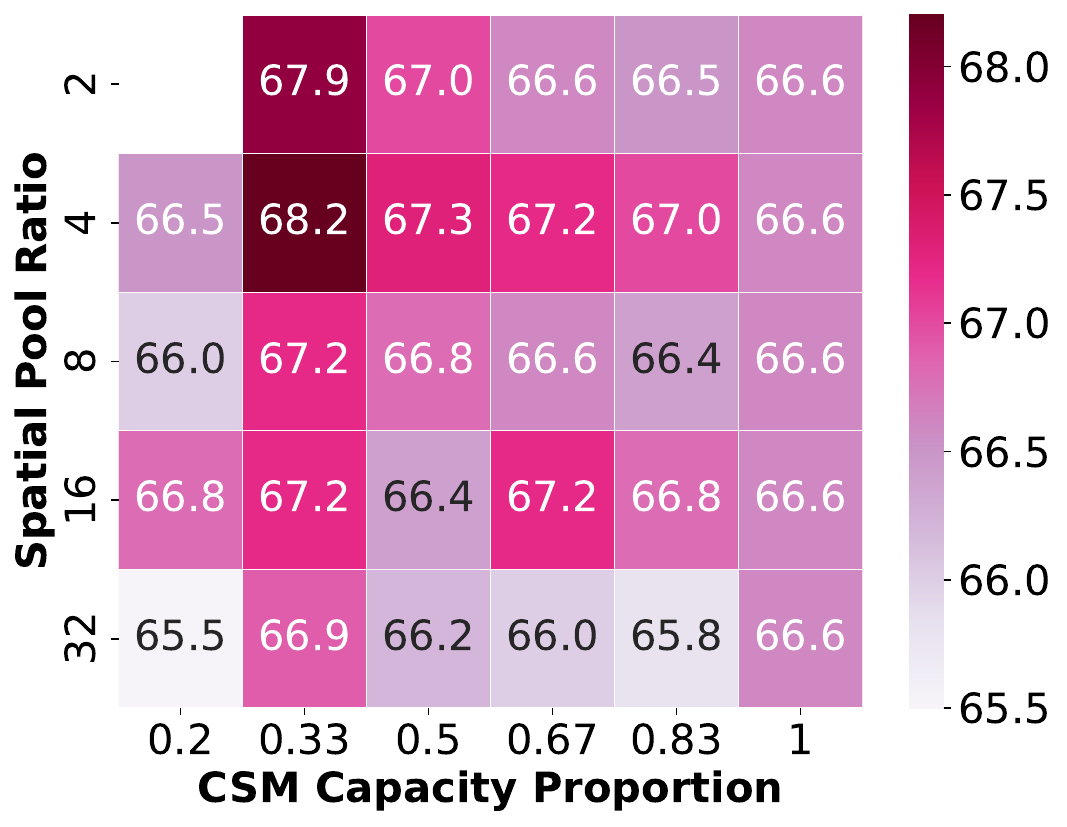}
        \caption{EgoSchema}
    \end{subfigure}
    \hfill
    \begin{subfigure}[b]{0.49\linewidth}
        \centering
        \includegraphics[width=\linewidth]{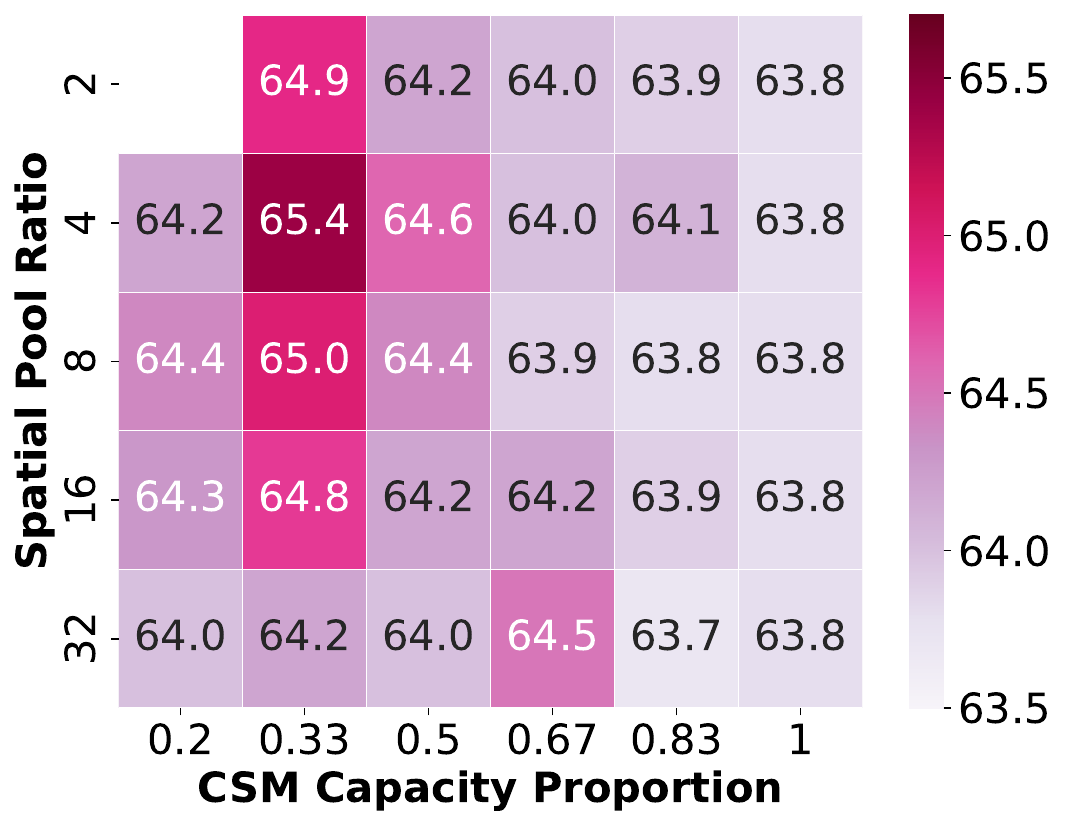}
        \caption{MVBench}
    \end{subfigure}
    \caption{
        \textbf{Impact of Pool Ratio} $R_{\text{pool}}$ \textbf{and CSM Capacity Proportion} $R_{\text{CSM}}$.
        The upper left grid is blank since its setting is invalid.
        % Experimentally, allocating one-third of total capacity to CSM when spatial pool ratio=4 yields the best performance.
        }
    \label{fig:capacity}
    \vspace{-5pt}
\end{figure}

%%%%%%%%%%% New
\begin{table*}[!t]
    \centering
    \small
    \setlength{\tabcolsep}{6.4pt}
    \begin{tabular}{lcccccccc}
        % \toprule
        \Xhline{0.8px}
            \rowcolor{lightblue} & & & & & & 
                \multicolumn{2}{c}{\textbf{Video-MME}}  \\
            \rowcolor{lightblue}
                \multirow{-2}{*}{\textbf{Model}} & 
                \multirow{-2}{*}{\textbf{Max \textit{N}\textsubscript{Vtokens}}} & 
                \multirow{-2}{*}{\textbf{EgoSchema}} & 
                \multirow{-2}{*}{\textbf{MLVU\textsubscript{dev}}} & 
                \multirow{-2}{*}{\textbf{LVBench}} &
                \multirow{-2}{*}{\textbf{MVBench}} & 
                \textbf{\textit{w/o subs}} & 
                \textbf{\textit{w/ subs}}  \\
        \hline
        \multicolumn{8}{l}{\textit{Offline video language models}} \\
        \hline
            MovieChat~\cite{moviechat} 
                & 32     & -    & -    & 22.5 & -    & -    & -    \\ 
            TimeChat~\cite{ren2024timechat} 
                & 96     & 33.0 & 30.9 & 22.3 & -    & -    & -    \\ 
            LLaMA-VID~\cite{llamavid} 
                & 2fps   & 38.5 & 33.2 & 23.9 & 41.9 & -    & -    \\ 
            ChatUni-Vi~\cite{chatunivi}     
                & 896    & -    & -    & -    & -    & 40.6 & 45.9 \\ 
            ShareGPT4-video~\cite{chen2024sharegpt4video} 
                & 9216   & -    & 46.4 & -    & 51.2 & 39.9 & 43.6 \\ 
            Video-Chat2-HD~\cite{li2024mvbench_videochat2} 
                & 512    & 55.8 & 47.9 & -    & 62.3 & 45.3 & 55.7 \\ 
            VideoLLaMA2~\cite{cheng2024videollama2} 
                & 9216   & 51.7 & 48.5 & -    & 54.6 & 47.9 & 50.3 \\ 
            LongVA~\cite{zhang2024longva}   
                & 18432  & -    & 56.3 & -    & -    & 52.6 & 54.3 \\ 
            LLaVA-OneVision~\cite{li2024llavaov} 
                & 6272   & 60.1 & 64.7 & -    & 56.7 & 58.2 & 61.5 \\ 
            LongVILA~\cite{xue2024longvila} 
                & 50176  & -    & -    & -    & -    & 57.5 & 61.8 \\ 
            Kangaroo~\cite{liu2024kangaroo} 
                & 16384  & 62.7 & 61.0 & -    & 61.0 & 56.0 & 57.6 \\ 
            Oryx-MLLM~\cite{liu2024oryx}    
                & 14400  & -    & -    & 30.4 & 63.9 & 58.3 & 62.6 \\ 
            % Qwen2-VL & 24576 & \underline{66.7} & \textbf{67.0} & \textbf{63.3} & \textbf{69.0} & - & - \\ 
            Qwen2-VL\textsuperscript{\textasteriskcentered}~\cite{wang2024qwen2vl}
                & 24576  & \underline{64.8} & \underline{66.0} & \underline{41.4} & \underline{65.1} & \underline{61.1} & \underline{65.9} \\ 
        \hline
        \multicolumn{2}{l}{\textit{Online video language models}} \\
        \hline
            VideoLLM-Online~\cite{chen2024videollmonline}
                & 2fps & 32.8 & 35.2 & 24.0 & 33.9 & 26.9 & 29.9 \\
            VideoLLM-MOD~\cite{wu2024videollmmod}
                & 1fps  & -       & -       & -       
                & -     & 49.2    & -       \\
            VideoStreaming~\cite{qian2024videostreaming} 
                & 256 & 44.1 & - & - & - & - & - \\
            Qwen2-VL-online\textsuperscript{\textdagger}
                & 11520 & 64.0 & 62.9 & 39.8 
                & 63.3  & 59.4 & 65.1 \\ 
            % \rowcolor{gray!30} Flash-VStream & 10800 & \textbf{68.4}  & \textbf{65.4} & \textbf{61.2} & \textbf{67.0} & \textbf{66.0} & \textbf{42.0} \\ 
            \rowcolor{gray!10} Flash-VStream (Ours) & 11520 
                & \textbf{68.2} (\textcolor{blue}{$\uparrow$ 4.2})
                & \textbf{66.3} (\textcolor{blue}{$\uparrow$ 3.4})
                & \textbf{42.0} (\textcolor{blue}{$\uparrow$ 2.2}) 
                & \textbf{65.4} (\textcolor{blue}{$\uparrow$ 2.1})
                & \textbf{61.2} (\textcolor{blue}{$\uparrow$ 1.8})
                & \textbf{67.0} (\textcolor{blue}{$\uparrow$ 1.9})
                \\ 
            % (\textcolor{blue}{$\uparrow$ 6.1})
        % \bottomrule
        \Xhline{0.8px}
    \end{tabular}
    \caption{\textbf{Comparison with state-of-the-art video language models on video question answering benchmarks.}
    We conduct experiments on five mainstream multiple-choice benchmarks.
    \textit{N}\textsubscript{Vtokens} denotes the number of video tokens used during evaluation. 
    \textit{subs} is short for subtitles.
    Qwen2-VL\textsuperscript{\textasteriskcentered} denotes the reproduced results under the official setting.
    Qwen2-VL-online\textsuperscript{\textdagger} denotes Qwen2-VL tested under real-time restriction (\textit{N}\textsubscript{Vtokens} <= 11520).
    The best two results are \textbf{bold-faced} and \underline{underlined}, respectively.
    }
    \label{tab:maintab}  
    \vspace{-10pt}
\end{table*}

\subsection{Adaptive Multimodal RoPE}
Rotary Position Embedding (RoPE)~\cite{su2024rope} is a widely used position embedding method in LLMs~\cite{touvron2023llama,Touvron2023Llama2O,dubey2024llama3}.
We improve upon the original Multimodal RoPE (M-RoPE)~\cite{wang2024qwen2vl,su2024rope} to support flexible positions, resulting in the Adaptive Multimodal RoPE (AM-RoPE). 
M-RoPE first splits the hidden dimension to three groups representing time, height and width axes, so that the position index becomes a triplet $(n_t,n_h,n_w)$. 
AM-RoPE is designed for flexible relative position embedding. The key is to use average positions to represent the compressed cluster feature.
In Flash-VStream, there are two types of video token, the CSM token and DAM token. 
For a DAM token at position $(x,y)$ of $M^{\text{DAM}}_{k}$, the AM-RoPE is calculated as: $n_t=P^{\text{DAM}}_k=t(k),n_h=y\times 2,n_w=x\times 2$ to accommodate the effect of average pooling. 
For a CSM token at position $(x,y)$ of $M^{\text{CSM}}_k$, $n_t=P^{\text{CSM}}_k,n_h=y,n_w=x$.

When a new question is posed, the question handler process begins to infer based on current Flash Memory 
$M^F$ containing 
$N_{\text{Vtokens}} = N^{\text{CSM}}\times h^\prime\times w^\prime + N^{\text{DAM}}\times h\times w$ tokens. By adjusting this budget, it is possible to ensure efficient real-time response in our asynchronous framework.
We conduct a speed test in~\Cref{sec:exp_efficiency} and find that a 7b model can achieve real-time inference by limiting 
$N_{\text{Vtokens}} \leq 12000$, as presented in~\Cref{fig:efficiency}.

\section{Experiments}

\subsection{Experimental setup}

\noindent\textbf{Implementation Details.}
Following Qwen2-VL, we employ a ViT with 3D patch embedding layer as the visual encoder, a merger projector as the projector.
Therefore, every two adjacent frames are temporally pooled before being encoded, and four adjacent ViT tokens are spatially pooled to one LLM token.
The visual encoder, projector and LLM are initialized from a pretrained MLLM, Qwen2-VL-7b~\cite{wang2024qwen2vl}.
\Cref{tab:config} shows detailed configurations of the proposed Flash Memory and input frames. 
As discussed in~\Cref{sec:exp_ablation}, we empirically assign 1/3 tokens to CSM and 2/3 tokens to DAM. 
The Flash Memory costs 11520 LLM tokens in total.

\noindent\textbf{Training Settings.}
To enhance video understanding ability based on Flash Memory, we adopt a LoRA~\cite{hu2022lora} instruction tuning stage. 
With the parameters of the visual encoder frozen, all linear layers of projector and LLM are LoRA finetuned.
We adopted a 9k subset from LLaVA-Video dataset~\cite{zhang2024llavavideo}, which contains instruction-following tasks like captioning, open-ended VQA and multiple-choice VQA.
More training details are provided in the supplementary.

\begin{table*}[!ht]
    \centering
    \small
    \begin{tabular}{lc|cccc|cccc}
        \Xhline{0.8px}
        \rowcolor{lightblue}
             &  &
            \multicolumn{4}{c|}{\textbf{\textit{w/o subtitles}}} & 
            \multicolumn{4}{c}{\textbf{\textit{w/ subtitles}}} \\ 
        \rowcolor{lightblue}
            \multirow{-2}{*}{\textbf{Model}} & 
            \multirow{-2}{*}{\textbf{Max \textit{N}\textsubscript{Vtokens}}} & 
             \textbf{Short} & \textbf{Middle} & \textbf{Long} & \textbf{Average} & 
             \textbf{Short} & \textbf{Middle} & \textbf{Long} & \textbf{Average} \\ 
        \hline
        ChatUni-Vi~\cite{chatunivi} & 
        896 & 45.7 & 40.3 & 35.8 & 40.6 & 51.2 & 44.6 & 41.8 & 45.9 \\ 
        ShareGPT4-video~\cite{chen2024sharegpt4video} & 
        9216 & 48.3 & 36.3 & 35.0 & 39.9 & 53.6 & 39.3 & 37.9 & 43.6 \\ 
        VideoLLaMA2~\cite{cheng2024videollama2} & 
        9216 & 56.0 & 45.4 & 42.0 & 47.9 & 59.4 & 47.6 & 43.8 & 50.3 \\ 
        LongVA~\cite{zhang2024longva} & 
        18432 & 61.1 & 50.4 & 46.2 & 52.6 & 61.1 & 53.6 & 47.6 & 54.3 \\ 
        LongVILA~\cite{xue2024longvila} & 
        147456 & 69.3 & 56.1 & 47.0 & 57.5 & 70.8 & 60.6 & 54.0 & 61.8 \\ 
        Kangaroo~\cite{liu2024kangaroo} & 
        16384 & 66.1 & 55.3 & 46.7 & 56.0 & 68.0 & 55.4 & 49.3 & 57.6 \\ 
        Qwen2-VL\textsuperscript{\textasteriskcentered}~\cite{wang2024qwen2vl} & 
        24576 & \underline{71.3} & \textbf{61.3} & \textbf{50.7} & \underline{61.1} & 71.1 & \textbf{69.0} & 57.6 & \underline{65.9} \\ 
        \hline
        Qwen2-VL-online\textsuperscript{\textdagger} & 
        11520 & 70.3 & 59.8 & 48.2 & 59.4 & \underline{71.4} & 66.0 & \underline{57.9} & 65.1 \\ 
        \rowcolor{gray!10} Flash-VStream (Ours)  & 11520 & \textbf{72.0} & \underline{61.1} & \underline{50.3} & \textbf{61.2} & \textbf{72.4} & \underline{67.0} & \textbf{61.4} & \textbf{67.0} \\ 
        \Xhline{0.8px}
    \end{tabular}
    \caption{\textbf{Comparison with state-of-the-art video language models on Video-MME benchmark.}
    \textit{N}\textsubscript{Vtokens} denotes the number of video tokens used during evaluation. 
    Qwen2-VL\textsuperscript{\textasteriskcentered} denotes the reproduced results under the official setting.
    Qwen2-VL-online\textsuperscript{\textdagger} denotes Qwen2-VL tested under real-time restriction (\textit{N}\textsubscript{Vtokens} <= 11520).
    The best two results are \textbf{bold-faced} and \underline{underlined}, respectively.
    }
    \label{tab:videomme}
    \vspace{-10pt}
\end{table*}

\noindent\textbf{Evaluation Settings.}
% It is worth noting that some previous works~\cite{moviechat,chatunivi,llamavid} follow~\cite{video-chatgpt} to test models on open-ended VQA benchmarks based on GPT-3.5 judgement. Since GPT APIs are proprietary and upgrade over time, this evaluation approach lacks reliability, stability and reproducibility~\cite{li2024seedbench}. 
For more reliable analysis, we conduct zero-shot multiple-choice VQA experiments on three long video benchmarks and two comprehensive video benchmarks. We report the multiple-choice accuracy on each benchmark.
EgoSchema~\cite{egoschema} is a long-form VQA dataset that is specifically designed for understanding first-person behaviors. 
MLVU~\cite{zhou2024mlvu} is a multi-task long video benchmark covering various video genres.
LVBench~\cite{wang2024lvbench} is an extreme long video understanding benchmark designed to test long-term comprehension capabilities of models.
MVBench~\cite{li2024mvbench_videochat2} contains amounts of temporal-related tasks, aiming at testing temporal understanding ability.
Video-MME~\cite{fu2024videomme} is a high-quality comprehensive video analysis benchmark, which contains videos ranging from 11 seconds to 1 hour.

Flash-VStream is evaluated on these benchmarks in an online setting. First, frames are extracted at 1 fps. Each frame is then fed to the Flash-VStream model sequentially, with the question posed at the end of the video frame stream.

\subsection{Computational Efficiency}
\label{sec:exp_efficiency}
We first measure the response latency of the Flash-VStream model by counting the response wall time of the question handler process under different cost limitations, as presented in \Cref{fig:efficiency}. Cost limitation confines the maximum number of video tokens during inference, namely \textit{N}\textsubscript{Vtokens}. All experiments are tested on a single A100 GPU with BFloat16 precision and FlashAttention-2~\cite{dao2024flashattention2}.
The experimental results show that our model can achieve real-time response with $N_{\text{Vtokens}} <= 12000$.
% Additionally, it is observed that the response latency exhibits a linear relationship with the sequence length. 
% This can be attributed to the infrastructural optimization of FlashAttention, which allows for linear time complexity with sequence length.
Models like Kangaroo~\cite{liu2024kangaroo}, Qwen2-VL~\cite{wang2024qwen2vl}, LongVA~\cite{zhang2024longva} and LongVILA~\cite{xue2024longvila} require far more than 12k tokens for inference, resulting in latency higher than 1 second. 
In contrast, Flash-VStream leverages a multi-process framework with an efficient memory design and
outperforms all competitive methods at the same token cost.

\subsection{Main Results}

In comparison to previous competitive models~\cite{moviechat,ren2024timechat,llamavid,chatunivi,chen2024sharegpt4video,li2024mvbench_videochat2,cheng2024videollama2,zhang2024longva,xue2024longvila,liu2024kangaroo,liu2024oryx,wang2024qwen2vl,chen2024videollmonline,qian2024videostreaming}
, Flash-VStream shows superior understanding capabilities on various challenging video understanding benchmarks, as detailed in~\Cref{tab:maintab}. 
For Qwen2-VL~\cite{wang2024qwen2vl}, we re-evaluate it under official setting and find it difficult to reproduce the reported results.
Although Flash-VStream performs slightly behind Qwen2-VL on MLVU dataset, our method is more efficient and saves 53\% of video tokens during inference. 
Flash-VStream significantly promotes the development of online VLMs, which are able to respond to human instructions in real time.

For a fair comparison, we slightly modify the testing setting of Qwen2-VL~\cite{wang2024qwen2vl} by limiting the maximum number of video tokens to 11520, resulting in a ``Qwen2-VL-online'' baseline.
As reported in~\Cref{tab:maintab}, Flash-VStream significantly surpasses Qwen2-VL-online on both short video and long video benchmarks under equal-cost condition, which demonstrates the effectiveness of the proposed Flash Memory.

\Cref{tab:videomme} provides a comprehensive comparison of model performance on the Video-MME benchmark. Leveraging the CSM and DAM memory, Flash-VStream achieves an effective trade-off between accuracy and efficiency across various video types.

\subsection{Ablation Study} 
\label{sec:exp_ablation}

\noindent\textbf{Flash Memory.}
We conduct an ablation study to evaluate the effectiveness of CSM and DAM in the Flash Memory. 
% As aforementioned, our method leverages a CSM module to aggregate long-context temporal information and a DAM module to retrieve detailed spatial information, aiming to reduce temporal redundancy under limited computation resources. 
From the results in~\Cref{tab:abl_flash_memory}, one can observe that both CSM and DAM bring significant and highly generalizable improvements across three benchmarks. 
There are three groups of experiments in~\Cref{tab:abl_flash_memory}. 
The first group evaluates the impact of removing components from the baseline setting row \ding{172}. CSM and DAM help improve the average accuracy by 2.0\% and 0.7\% compared to uniform sampling, respectively.
The second group further evaluates the influence of CSM with the existence of uniformly sampled DAM in row \ding{176} (since full DAM relies on full CSM). 
The third group compares the implementation details of Qwen2-VL-online setting in~\Cref{tab:maintab}, which show different preferences for each benchmark. 

\begin{table*}[t]
    \centering
    \footnotesize
    \begin{tabular}{c|ccccc|cccc}
        \Xhline{0.8px}
            \rowcolor{lightblue}
                & \multicolumn{5}{c|}{\textbf{Memory Component Settings}} & 
                \multicolumn{4}{c}{\textbf{Evaluation Results}} \\
            \rowcolor{lightblue}
                \multirow{-2}{*}{\textbf{ID}} & \textbf{CSM} & \textbf{DAM} & 
                \textbf{N\textsubscript{Vtokens}} & \textbf{CSM Size} & \textbf{DAM Size} & 
                \textbf{MVBench} & \textbf{EgoSchema} & \textbf{Video-MME(w/o)} & \textbf{Average} \\
        \hline
            \rowcolor{gray!10}
            \ding{172} & \ding{51} & \ding{51} & 11520 & $60 \times 64$ & $30 \times 256$ & 65.4 & 68.2 & 61.2 & 64.9 \\
            \ding{173} & \ding{51} & Uni. Smp. & 11520 & $60 \times 64$ & $30 \times 256$ & 64.3 & 67.8 & 60.6 & 64.2 \\
            \ding{174} & \ding{51} & \ding{55} & 3840  & $60 \times 64$ & $0$ & 64.0 & 66.8 & 60.1 & 63.6 \\
            \ding{175} & Uni. Smp. & \ding{55} & 3840  & $60 \times 64$ & $0$ & 62.4 & 63.4 & 59.0 & 61.6 \\
        \hline
            \rowcolor{gray!10}
            \ding{176} & \ding{51} & Uni. Smp. & 11520 & $60 \times 64$ & $30 \times 256$ & 64.3 & 67.8 & 60.6 & 64.2 \\
            \ding{177} & Uni. Smp. & Uni. Smp. & 11520 & $60 \times 64$ & $30 \times 256$ & 63.8 & 66.0 & 59.6 & 63.1 \\
            \ding{178} & \ding{55} & Uni. Smp. & 7680  & $0$ & $30 \times 256$ & 63.1 & 65.7 & 59.3 & 62.7 \\
        \hline
            \ding{179} & Uni. Smp. & \ding{55} & 11520 & $180 \times 64$ & $0$ & 63.3 & 64.0 & 59.4 & 62.3 \\
            \ding{180} & \ding{55} & Uni. Smp. & 11520 & $0$ & $45 \times 256$ & 63.2 & 65.1 & 59.0 & 62.4 \\
        \Xhline{0.8px}
    \end{tabular}
    \caption{\textbf{Analysis of the design of Flash Memory.} 
    We investigate the effects of eliminating the two components of Flash Memory: CSM and DAM.
    ``Uni. Smp.'' stands for ``Uniform Sample''. 
    The first row is the default baseline setting of our model.
    The penultimate row is the same as the Qwen2-VL-online setting in~\Cref{tab:maintab}.
    }
    \label{tab:abl_flash_memory}  
    \vspace{-10pt}
\end{table*}

\noindent\textbf{Memory Capacity Allocation.}
We investigate the capacity allocation strategy for CSM and DAM with a fixed total memory tokens budget.
For \textbf{real-time inference}, the amount of visual tokens should not exceed 12000 (\Cref{fig:efficiency}).
Under this fixed budget condition, we adjust the proportion of CSM capacity in total memory 
$R_\text{CSM}=S^{\text{CSM}} / (S^{\text{CSM}}+S^{\text{DAM}})$ 
by controlling $N^{\text{CSM}}$ and $N^{\text{DAM}}$;
and adjust the pool ratio $R_\text{pool}=(h\times w)/(h^\prime\times w^\prime)$
by controlling $h\times w$ and $N^{\text{DAM}}$.
The grid search result in~\Cref{fig:capacity} shows that allocating around one-third of the total memory capacity to CSM when pool ratio = 4 yields the best performance for EgoSchema and MVBench. 
This suggests that a balanced allocation strategy, where CSM is given sufficient capacity to capture long-term temporal information while DAM retains enough capacity to preserve detailed spatial information, is crucial for optimal performance.
More ablation studies on memory structure configuration can be found in the supplementary.

\begin{table}[t]
    \centering
    \footnotesize
    \setlength{\tabcolsep}{8pt}
    \begin{tabular}{c|l|ccc}
        \Xhline{0.8px}
        \rowcolor{lightblue}
            \textbf{ID} & \textbf{Clustering Policy} & \textbf{MVB} & \textbf{EGO} & \textbf{MME} \\ 
        \hline
            \rowcolor{gray!10} 
            \ding{174} & K-means~\cite{macqueen1967some}& 64.0 & 66.8 & 60.1 \\ 
            & DBScan~\cite{ester1996dbscan}             & 63.8 & 66.6 & 59.7 \\ 
            & GMM                                       & 59.6 & 65.8 & 59.5 \\ 
            & Neighbor Merge~\cite{moviechat}           & 63.7 & 65.0 & 59.4 \\ 
            & Neighbor Drop                             & 63.8 & 62.4 & 59.4 \\ 
            \ding{175} & Uniform Sample                 & 62.4 & 63.4 & 59.0 \\ 
        \Xhline{0.8px}
    \end{tabular}
    \caption{\textbf{Ablation study on CSM clustering policy.} 
    This experiment compares K-means with other clustering methods. DAM is removed to isolate the effect of CSM. 
    }
    \label{tab:abl_csm}    
    \vspace{-10pt}
\end{table}
\noindent\textbf{CSM Clustering Policy.}
In~\Cref{tab:abl_csm}, we compare the K-means clustering algorithm with other clustering methods.
We remove DAM in Flash Memory to isolate the effect of CSM and take row \ding{174} in~\Cref{tab:abl_flash_memory} as a baseline.
For example, DBScan~\cite{ester1996dbscan} is robust to noise and outliers. 
GMM (Gaussian Mixture Model) is a clustering method based on probability distributions, which is appropriate for statistical analysis. 
MovieChat~\cite{moviechat} utilizes a ``Neighbor Merge'' method for memory updating. It merges the tokens of the two most similar adjacent frames at every step.
To examine the effect of token merge, we implement a similar method named ``Neighbor Drop'', which randomly chooses a frame to keep and another to drop, instead of merging them. All the methods above are compared to uniform sampling.

As presented in~\Cref{tab:abl_csm}, K-means outperforms other clustering methods in the EgoSchema and MVBench datasets, surpassing the uniform sampling baseline by 3.4\% and 1.6\%, respectively. 
DBScan, while robust to noise and outliers, slightly underperforms K-means. We attribute this to the fact that this is a cluster number known task, where DBScan may not be the best choice.
Note that although Neighbor Merge achieves equally good accuracy on MVBench, it falls behind in long video understanding tasks in EgoSchema.

\noindent\textbf{DAM Sampling Policy.}
DAM is designed for storing details of the most informative key frames.
As proposed in~\Cref{sec:dam}, the default Feature-Centric retrieval method relates the importance of a frame to the distance between its feature map and the cluster centroid feature map. 
It selects $N^{\text{DAM}}$ frames nearest to the centroids of the top-$N^{\text{DAM}}$ largest clusters as DAM.
% Furthermore, additional sampling methods are included for comprehensive ablation. 
We compare it with other methods from~\Cref{tab:abl_flash_memory} row \ding{172} baseline.
Temporal-Centric retrieval selects frames nearest to the temporal position $P^{\text{CSM}}_i$ of top-$N^{\text{DAM}}$ largest clusters. 
Cosine Similarity method replaces the Euclidean distance in~\Cref{eq:dist} with cosine similarity.
As presented in~\Cref{tab:abl_dam}, Feature-Centric retrieval demonstrates robust improvement on short video and long video understanding benchmarks. Cosine Similarity results in relatively lower performance, indicating the magnitude of features matters.

As shown in the lower part of~\Cref{tab:abl_dam}, top-k largest selection method outperforms top-k smallest method and uniform-k method, being comparable to sampling all frames.

\begin{table}[t]
    \centering
    \footnotesize    
    \setlength{\tabcolsep}{5pt}
    \begin{tabular}{c|ll|ccc}
        \Xhline{0.8px}
        \rowcolor{lightblue}
            \textbf{ID} & \textbf{Retrieval Policy} & \textbf{Selection Policy} & \textbf{MVB} & \textbf{EGO} & \textbf{MME} \\ 
        \hline
            \rowcolor{gray!10} 
            \ding{172} & Feature-Centric   & Top-k largest  & 65.4 & 68.2 & 61.2 \\ 
            & Temporal-Centric  & Top-k largest  & 64.4 & 67.8 & 60.8 \\
            & Cosine Similarity & Top-k largest  & 64.1 & 66.5 & 60.2 \\
            \ding{173} & Uniform Sample    & k frames       & 64.3 & 67.8 & 60.6 \\ 
        \hline
            \rowcolor{gray!10} 
            \ding{172} & Feature-Centric   & Top-k largest  & 65.4 & 68.2 & 61.2 \\ 
            & Feature-Centric   & Top-k smallest & 63.9 & 67.6 & 59.6 \\ 
            & Uniform Sample    & All frames     & 65.5 & 68.6 & 61.2 \\ 
        \Xhline{0.8px}
    \end{tabular}
    \caption{\textbf{Ablation study on DAM sampling policy.} 
    This experiment compares different retrieval methods and selection methods. 
    Here $k=N^{\text{DAM}}$ and the last row uses 19200 visual tokens.
    }
    \label{tab:abl_dam}
    \vspace{-10pt}
\end{table}

\subsection{Memory Visualization and Case Study}

% main method
\begin{figure*}[t]
    \centering
    \includegraphics[width=1\linewidth,height=0.53\textheight]{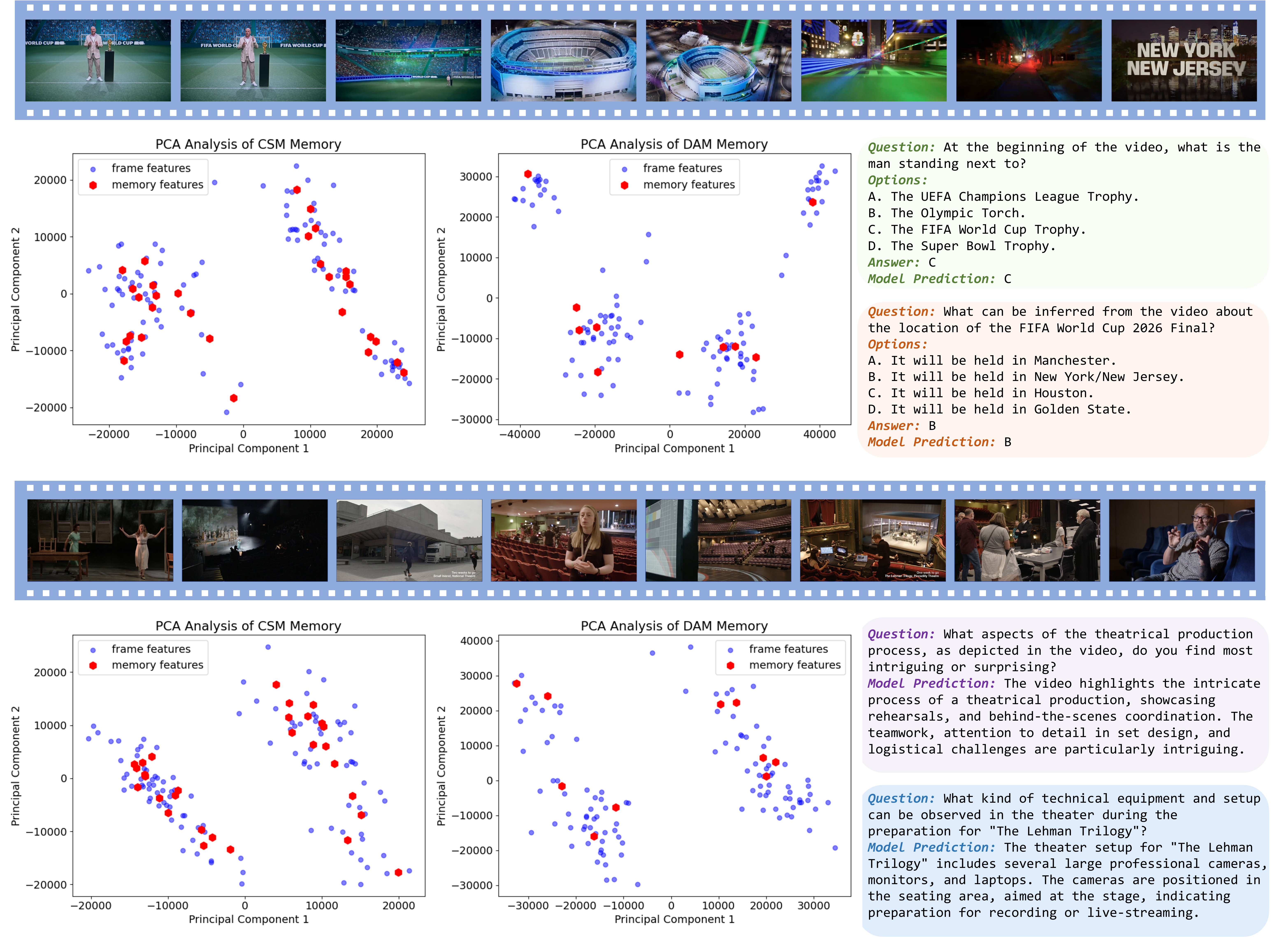}
    \caption{\textbf{Memory Distribution Visualization and Case Study.} The left side presents a PCA visualization of the Flash Memory distribution in the feature space. Each point in it stands for a feature map of a single frame or a slice of memory. The CSM and DAM appropriately represent the distributional characteristics of the feature clusters.
    On the right side, different types of question answering cases show exceptional proficiency of the Flash-VStream model.
    (upper: multiple-choice questions; lower: open-ended questions)
    }
    \vspace{-5pt}
    \label{fig:case} 
\end{figure*}

We investigate the memory consolidation procedure in the deep feature space by dimension reduction with Principal Component Analysis (PCA).
As illustrated on the left side of~\Cref{fig:case}, each red hexagon stands for a memory token of CSM or DAM, and each blue point stands for a frame feature map ($e^{\text{L}}_t$ in comparison with CSM and $e^{\text{H}}_t$ in comparison with DAM).
The PCA plots show that the memory features are distinctly separated from the frame features, suggesting that the CSM and DAM mechanisms are successful in encoding and consolidating important information.
% ---- ----
We note that CSM and DAM formulate different shapes of clusters for each video. For example, in the upper part of~\Cref{fig:case}, DAM has four densely distributed areas while CSM has only two. We suppose that this is because high resolution features have more discriminability.
The visualization proves that the Flash Memory features effectively capture the underlying structure of the feature space.

% \subsection{Case Study}
On the right side of~\Cref{fig:case}, we present several question-answering cases that illustrate the efficacy of the Flash-VStream model. 
The case study encompasses both multiple-choice questions and open-ended questions, including those related to spatial cognition, optical character recognition (OCR), and complex reasoning.
These results demonstrate that Flash-VStream not only excels in understanding and processing long video content but also in providing precise and contextually relevant answers, thereby validating its practical applicability in real-world scenarios.
For more case studies, please refer to the supplementary materials.

\section{Conclusion}
% \vspace{-20pt}

In conclusion, we introduce Flash-VStream, an efficient video-language model for understanding long video streams and providing real-time responses to user queries.
Flash-VStream uses a two-process asynchronous framework to separate vision and language processing, ensuring real-time responses.
The core innovation lies in the Flash Memory module, which comprises a Context Synopsis Memory for long-term temporal information aggregation and a Detail Augmentation Memory for retrieving detailed spatial information. 
This design is based on the observation that temporal redundancy is prevalent in videos.
Extensive experiments on multiple comprehensive video benchmarks demonstrate the superior performance and efficiency of Flash-VStream compared to existing state-of-the-art models. 
We hope our work will inspire further research and advancements in the field of efficient long video understanding.

\section*{Acknowledgements}

This work was supported in part by the National Key Research and Development Program of China under Grant 2023YFF1105101,
and in part by the Guangdong Natural Science Funds for Distinguished Young Scholar (No. 2025B1515020012).

\clearpage

    {
        \small
        \bibliographystyle{ieeenat_fullname}
        \bibliography{main}

\begin{thebibliography}{84}
\providecommand{\natexlab}[1]{#1}
\providecommand{\url}[1]{\texttt{#1}}
\expandafter\ifx\csname urlstyle\endcsname\relax
  \providecommand{\doi}[1]{doi: #1}\else
  \providecommand{\doi}{doi: \begingroup \urlstyle{rm}\Url}\fi

\bibitem[Achiam et~al.(2023)Achiam, Adler, Agarwal, Ahmad, Akkaya, Aleman, Almeida, Altenschmidt, Altman, Anadkat, et~al.]{achiam2023gpt4report}
Josh Achiam, Steven Adler, Sandhini Agarwal, Lama Ahmad, Ilge Akkaya, Florencia~Leoni Aleman, Diogo Almeida, Janko Altenschmidt, Sam Altman, Shyamal Anadkat, et~al.
\newblock Gpt-4 technical report.
\newblock \emph{arXiv preprint arXiv:2303.08774}, 2023.

\bibitem[Bai et~al.(2024)Bai, Liu, Han, Zhang, and Tang]{bai2024self}
Sule Bai, Yong Liu, Yifei Han, Haoji Zhang, and Yansong Tang.
\newblock Self-calibrated clip for training-free open-vocabulary segmentation.
\newblock \emph{arXiv preprint arXiv:2411.15869}, 2024.

\bibitem[Balazevic et~al.(2024)Balazevic, Shi, Papalampidi, Chaabouni, Koppula, and H{\'e}naff]{balavzevic2024mcvit}
Ivana Balazevic, Yuge Shi, Pinelopi Papalampidi, Rahma Chaabouni, Skanda Koppula, and Olivier~J H{\'e}naff.
\newblock Memory consolidation enables long-context video understanding.
\newblock In \emph{ICML}, pages 2527--2542, 2024.

\bibitem[Brown et~al.(2020)Brown, Mann, Ryder, Subbiah, Kaplan, Dhariwal, Neelakantan, Shyam, Sastry, Askell, et~al.]{brown2020language}
Tom Brown, Benjamin Mann, Nick Ryder, Melanie Subbiah, Jared~D Kaplan, Prafulla Dhariwal, Arvind Neelakantan, Pranav Shyam, Girish Sastry, Amanda Askell, et~al.
\newblock Language models are few-shot learners.
\newblock \emph{NeurIPS}, pages 33, 1877--1901, 2020.

\bibitem[Chang et~al.(2018)Chang, Sun, Wu, and Lin]{chang2018memory}
Yen-Yu Chang, Fan-Yun Sun, Yueh-Hua Wu, and Shou-De Lin.
\newblock A memory-network based solution for multivariate time-series forecasting.
\newblock \emph{arXiv preprint arXiv:1809.02105}, 2018.

\bibitem[Chen et~al.(2019)Chen, Li, Deng, Li, and Philip]{chen2019distributed}
Jianguo Chen, Kenli Li, Qingying Deng, Keqin Li, and S~Yu Philip.
\newblock Distributed deep learning model for intelligent video surveillance systems with edge computing.
\newblock \emph{IEEE Transactions on Industrial Informatics}, 2019.

\bibitem[Chen et~al.(2024)Chen, Lv, Wu, Lin, Song, Gao, Liu, Gao, Mao, and Shou]{chen2024videollmonline}
Joya Chen, Zhaoyang Lv, Shiwei Wu, Kevin~Qinghong Lin, Chenan Song, Difei Gao, Jia-Wei Liu, Ziteng Gao, Dongxing Mao, and Mike~Zheng Shou.
\newblock Videollm-online: Online video large language model for streaming video.
\newblock In \emph{CVPR}, pages 18407--18418, 2024.

\bibitem[Chen et~al.(2025)Chen, Wei, Li, Dong, Zhang, Zang, Chen, Duan, Tang, Yuan, et~al.]{chen2024sharegpt4video}
Lin Chen, Xilin Wei, Jinsong Li, Xiaoyi Dong, Pan Zhang, Yuhang Zang, Zehui Chen, Haodong Duan, Zhenyu Tang, Li Yuan, et~al.
\newblock Sharegpt4video: Improving video understanding and generation with better captions.
\newblock \emph{NeurIPS}, 37:\penalty0 19472--19495, 2025.

\bibitem[Cheng and Schwing(2022)]{cheng2022xmem}
Ho~Kei Cheng and Alexander~G Schwing.
\newblock Xmem: Long-term video object segmentation with an atkinson-shiffrin memory model.
\newblock In \emph{ECCV}, pages 640--658. Springer, 2022.

\bibitem[Cheng et~al.(2024)Cheng, Leng, Zhang, Xin, Li, Chen, Zhu, Zhang, Luo, Zhao, et~al.]{cheng2024videollama2}
Zesen Cheng, Sicong Leng, Hang Zhang, Yifei Xin, Xin Li, Guanzheng Chen, Yongxin Zhu, Wenqi Zhang, Ziyang Luo, Deli Zhao, et~al.
\newblock Videollama 2: Advancing spatial-temporal modeling and audio understanding in video-llms.
\newblock \emph{arXiv preprint arXiv:2406.07476}, 2024.

\bibitem[Dai et~al.(2024)Dai, Li, LI, Tiong, Zhao, Wang, Li, Fung, and Hoi]{Dai2023InstructBLIPTG}
Wenliang Dai, Junnan Li, DONGXU LI, Anthony Meng~Huat Tiong, Junqi Zhao, Weisheng Wang, Boyang Li, Pascale~N Fung, and Steven Hoi.
\newblock Instructblip: Towards general-purpose vision-language models with instruction tuning.
\newblock \emph{NeurIPS}, 36, 2024.

\bibitem[Dao(2024)]{dao2024flashattention2}
Tri Dao.
\newblock Flashattention-2: Faster attention with better parallelism and work partitioning.
\newblock In \emph{ICLR}, 2024.

\bibitem[Dosovitskiy et~al.(2021)Dosovitskiy, Beyer, Kolesnikov, Weissenborn, Zhai, Unterthiner, Dehghani, Minderer, Heigold, Gelly, Uszkoreit, and Houlsby]{dosovitskiy2020image}
Alexey Dosovitskiy, Lucas Beyer, Alexander Kolesnikov, Dirk Weissenborn, Xiaohua Zhai, Thomas Unterthiner, Mostafa Dehghani, Matthias Minderer, Georg Heigold, Sylvain Gelly, Jakob Uszkoreit, and Neil Houlsby.
\newblock An image is worth 16x16 words: Transformers for image recognition at scale.
\newblock In \emph{ICLR}, 2021.

\bibitem[Dubey et~al.(2024)Dubey, Jauhri, Pandey, Kadian, Al-Dahle, Letman, Mathur, Schelten, Yang, Fan, et~al.]{dubey2024llama3}
Abhimanyu Dubey, Abhinav Jauhri, Abhinav Pandey, Abhishek Kadian, Ahmad Al-Dahle, Aiesha Letman, Akhil Mathur, Alan Schelten, Amy Yang, Angela Fan, et~al.
\newblock The llama 3 herd of models.
\newblock \emph{arXiv preprint arXiv:2407.21783}, 2024.

\bibitem[Ester et~al.(1996)Ester, Kriegel, Sander, Xu, et~al.]{ester1996dbscan}
Martin Ester, Hans-Peter Kriegel, J{\"o}rg Sander, Xiaowei Xu, et~al.
\newblock A density-based algorithm for discovering clusters in large spatial databases with noise.
\newblock In \emph{KDD}, pages 226--231, 1996.

\bibitem[Fu et~al.(2025)Fu, Dai, Luo, Li, Ren, Zhang, Wang, Zhou, Shen, Zhang, et~al.]{fu2024videomme}
Chaoyou Fu, Yuhan Dai, Yongdong Luo, Lei Li, Shuhuai Ren, Renrui Zhang, Zihan Wang, Chenyu Zhou, Yunhang Shen, Mengdan Zhang, et~al.
\newblock Video-mme: The first-ever comprehensive evaluation benchmark of multi-modal llms in video analysis.
\newblock In \emph{CVPR}, pages 24108--24118, 2025.

\bibitem[Gan et~al.(2023)Gan, Wang, Sun, Wu, Guo, and Nie]{gan2023temporal}
Tian Gan, Xiao Wang, Yan Sun, Jianlong Wu, Qingpei Guo, and Liqiang Nie.
\newblock Temporal sentence grounding in streaming videos.
\newblock In \emph{ACM MM}, pages 4637--4646, 2023.

\bibitem[Gao et~al.(2023)Gao, Zhou, Ji, Zhu, Yang, and Shou]{gao2023mist}
Difei Gao, Luowei Zhou, Lei Ji, Linchao Zhu, Yi Yang, and Mike~Zheng Shou.
\newblock Mist: Multi-modal iterative spatial-temporal transformer for long-form video question answering.
\newblock In \emph{CVPR}, pages 14773--14783, 2023.

\bibitem[Gao et~al.(2024)Gao, Geng, Zhang, Ma, Fang, Zhang, Li, and Qiao]{gao2021clip}
Peng Gao, Shijie Geng, Renrui Zhang, Teli Ma, Rongyao Fang, Yongfeng Zhang, Hongsheng Li, and Yu Qiao.
\newblock Clip-adapter: Better vision-language models with feature adapters.
\newblock \emph{IJCV}, 132\penalty0 (2):\penalty0 581--595, 2024.

\bibitem[Ghodrati et~al.(2021)Ghodrati, Bejnordi, and Habibian]{ghodrati2021frameexit}
Amir Ghodrati, Babak~Ehteshami Bejnordi, and Amirhossein Habibian.
\newblock Frameexit: Conditional early exiting for efficient video recognition.
\newblock In \emph{CVPR}, pages 15608--15618, 2021.

\bibitem[Han et~al.(2021)Han, Huang, Song, Yang, Wang, and Wang]{han2021dynamicsurvey}
Yizeng Han, Gao Huang, Shiji Song, Le Yang, Honghui Wang, and Yulin Wang.
\newblock Dynamic neural networks: A survey.
\newblock \emph{TPAMI}, 44\penalty0 (11):\penalty0 7436--7456, 2021.

\bibitem[He et~al.(2018)He, Luo, Tian, and Zeng]{track3}
Anfeng He, Chong Luo, Xinmei Tian, and Wenjun Zeng.
\newblock A twofold siamese network for real-time object tracking.
\newblock In \emph{CVPR}, pages 4834--4843, 2018.

\bibitem[Hu et~al.(2022)Hu, yelong shen, Wallis, Allen-Zhu, Li, Wang, Wang, and Chen]{hu2022lora}
Edward~J Hu, yelong shen, Phillip Wallis, Zeyuan Allen-Zhu, Yuanzhi Li, Shean Wang, Lu Wang, and Weizhu Chen.
\newblock Lo{RA}: Low-rank adaptation of large language models.
\newblock In \emph{ICLR}, 2022.

\bibitem[Huang et~al.(2020)Huang, Xiong, Rao, Wang, and Lin]{huang2020movienet}
Qingqiu Huang, Yu Xiong, Anyi Rao, Jiaze Wang, and Dahua Lin.
\newblock Movienet: A holistic dataset for movie understanding.
\newblock In \emph{ECCV}, pages 709--727, 2020.

\bibitem[Jin et~al.(2024)Jin, Takanobu, Zhang, Cao, and Yuan]{chatunivi}
Peng Jin, Ryuichi Takanobu, Wancai Zhang, Xiaochun Cao, and Li Yuan.
\newblock Chat-univi: Unified visual representation empowers large language models with image and video understanding.
\newblock In \emph{CVPR}, pages 13700--13710, 2024.

\bibitem[Li et~al.(2024{\natexlab{a}})Li, Ge, Ge, Wang, Wang, Zhang, and Shan]{li2024seedbench}
Bohao Li, Yuying Ge, Yixiao Ge, Guangzhi Wang, Rui Wang, Ruimao Zhang, and Ying Shan.
\newblock Seed-bench: Benchmarking multimodal large language models.
\newblock In \emph{CVPR}, pages 13299--13308, 2024{\natexlab{a}}.

\bibitem[Li et~al.(2024{\natexlab{b}})Li, Zhang, Guo, Zhang, Li, Zhang, Zhang, Li, Liu, and Li]{li2024llavaov}
Bo Li, Yuanhan Zhang, Dong Guo, Renrui Zhang, Feng Li, Hao Zhang, Kaichen Zhang, Yanwei Li, Ziwei Liu, and Chunyuan Li.
\newblock Llava-onevision: Easy visual task transfer.
\newblock \emph{arXiv preprint arXiv:2408.03326}, 2024{\natexlab{b}}.

\bibitem[Li et~al.(2022)Li, Li, Xiong, and Hoi]{Li2022BLIPBL}
Junnan Li, Dongxu Li, Caiming Xiong, and Steven Hoi.
\newblock Blip: Bootstrapping language-image pre-training for unified vision-language understanding and generation.
\newblock In \emph{ICML}, pages 12888--12900, 2022.

\bibitem[Li et~al.(2023)Li, Li, Savarese, and Hoi]{li2023blip}
Junnan Li, Dongxu Li, Silvio Savarese, and Steven Hoi.
\newblock Blip-2: Bootstrapping language-image pre-training with frozen image encoders and large language models.
\newblock In \emph{ICML}, pages 19730--19742, 2023.

\bibitem[Li et~al.(2024{\natexlab{c}})Li, Wang, He, Li, Wang, Liu, Wang, Xu, Chen, Luo, et~al.]{li2024mvbench_videochat2}
Kunchang Li, Yali Wang, Yinan He, Yizhuo Li, Yi Wang, Yi Liu, Zun Wang, Jilan Xu, Guo Chen, Ping Luo, et~al.
\newblock Mvbench: A comprehensive multi-modal video understanding benchmark.
\newblock In \emph{CVPR}, pages 22195--22206, 2024{\natexlab{c}}.

\bibitem[Li et~al.(2025)Li, Wang, and Jia]{llamavid}
Yanwei Li, Chengyao Wang, and Jiaya Jia.
\newblock Llama-vid: An image is worth 2 tokens in large language models.
\newblock In \emph{ECCV}, pages 323--340. Springer, 2025.

\bibitem[Liu et~al.(2023)Liu, Li, Wu, and Lee]{llava}
Haotian Liu, Chunyuan Li, Qingyang Wu, and Yong~Jae Lee.
\newblock Visual instruction tuning.
\newblock \emph{NeurIPS}, 36:\penalty0 34892--34916, 2023.

\bibitem[Liu et~al.(2024{\natexlab{a}})Liu, Li, Li, and Lee]{Liu2023ImprovedBW}
Haotian Liu, Chunyuan Li, Yuheng Li, and Yong~Jae Lee.
\newblock Improved baselines with visual instruction tuning.
\newblock In \emph{CVPR}, pages 26296--26306, 2024{\natexlab{a}}.

\bibitem[Liu et~al.(2024{\natexlab{b}})Liu, Wang, Ma, Wu, Ma, Wei, Jiao, Wu, and Hu]{liu2024kangaroo}
Jiajun Liu, Yibing Wang, Hanghang Ma, Xiaoping Wu, Xiaoqi Ma, Xiaoming Wei, Jianbin Jiao, Enhua Wu, and Jie Hu.
\newblock Kangaroo: A powerful video-language model supporting long-context video input.
\newblock \emph{arXiv preprint arXiv:2408.15542}, 2024{\natexlab{b}}.

\bibitem[Liu et~al.(2022)Liu, Yu, Yin, Zhao, Zhao, Xia, and Yang]{track2}
Yong Liu, Ran Yu, Fei Yin, Xinyuan Zhao, Wei Zhao, Weihao Xia, and Yujiu Yang.
\newblock Learning quality-aware dynamic memory for video object segmentation.
\newblock In \emph{ECCV}, pages 468--486, 2022.

\bibitem[Liu et~al.(2024{\natexlab{c}})Liu, Zhang, Wang, Wang, Yang, and Tang]{multimodal3}
Yong Liu, Cairong Zhang, Yitong Wang, Jiahao Wang, Yujiu Yang, and Yansong Tang.
\newblock Universal segmentation at arbitrary granularity with language instruction.
\newblock In \emph{CVPR}, pages 3459--3469, 2024{\natexlab{c}}.

\bibitem[Liu et~al.(2024{\natexlab{d}})Liu, Dong, Liu, Hu, Lu, and Rao]{liu2024oryx}
Zuyan Liu, Yuhao Dong, Ziwei Liu, Winston Hu, Jiwen Lu, and Yongming Rao.
\newblock Oryx mllm: On-demand spatial-temporal understanding at arbitrary resolution.
\newblock \emph{arXiv preprint arXiv:2409.12961}, 2024{\natexlab{d}}.

\bibitem[Luo et~al.(2023)Luo, Xiao, Liu, Li, Wang, Tang, Li, and Yang]{luo2023soc}
Zhuoyan Luo, Yicheng Xiao, Yong Liu, Shuyan Li, Yitong Wang, Yansong Tang, Xiu Li, and Yujiu Yang.
\newblock Soc: Semantic-assisted object cluster for referring video object segmentation.
\newblock \emph{NeurIPS}, pages 26425--26437, 2023.

\bibitem[Luvizon et~al.(2020)Luvizon, Picard, and Tabia]{action2}
Diogo~C Luvizon, David Picard, and Hedi Tabia.
\newblock Multi-task deep learning for real-time 3d human pose estimation and action recognition.
\newblock \emph{IEEE TPAMI}, 43\penalty0 (8):\penalty0 2752--2764, 2020.

\bibitem[Ma et~al.(2024)Ma, Jin, Wang, Xian, Feng, and Yang]{vistallama}
Fan Ma, Xiaojie Jin, Heng Wang, Yuchen Xian, Jiashi Feng, and Yi Yang.
\newblock Vista-llama: Reducing hallucination in video language models via equal distance to visual tokens.
\newblock In \emph{CVPR}, pages 13151--13160, 2024.

\bibitem[Maaz et~al.(2024)Maaz, Rasheed, Khan, and Khan]{video-chatgpt}
Muhammad Maaz, Hanoona Rasheed, Salman Khan, and Fahad Khan.
\newblock Video-chatgpt: Towards detailed video understanding via large vision and language models.
\newblock In \emph{ACL}, pages 12585--12602, 2024.

\bibitem[MacQueen(1967)]{macqueen1967some}
J MacQueen.
\newblock Some methods for classification and analysis of multivariate observations.
\newblock In \emph{Proceedings of 5-th Berkeley Symposium on Mathematical Statistics and Probability/University of California Press}, 1967.

\bibitem[Mangalam et~al.(2023)Mangalam, Akshulakov, and Malik]{egoschema}
Karttikeya Mangalam, Raiymbek Akshulakov, and Jitendra Malik.
\newblock Egoschema: A diagnostic benchmark for very long-form video language understanding.
\newblock \emph{NeurIPS}, 36:\penalty0 46212--46244, 2023.

\bibitem[Muhammad et~al.(2019)Muhammad, Hussain, Del~Ser, Palade, and De~Albuquerque]{muhammad2019deepres}
Khan Muhammad, Tanveer Hussain, Javier Del~Ser, Vasile Palade, and Victor Hugo~C De~Albuquerque.
\newblock Deepres: A deep learning-based video summarization strategy for resource-constrained industrial surveillance scenarios.
\newblock \emph{IEEE Transactions on Industrial Informatics}, 16\penalty0 (9):\penalty0 5938--5947, 2019.

\bibitem[Ouyang et~al.(2022)Ouyang, Wu, Jiang, Almeida, Wainwright, Mishkin, Zhang, Agarwal, Slama, Ray, et~al.]{Ouyang2022TrainingLM}
Long Ouyang, Jeffrey Wu, Xu Jiang, Diogo Almeida, Carroll Wainwright, Pamela Mishkin, Chong Zhang, Sandhini Agarwal, Katarina Slama, Alex Ray, et~al.
\newblock Training language models to follow instructions with human feedback.
\newblock \emph{NeurIPS}, pages 27730--27744, 2022.

\bibitem[Qian et~al.(2025)Qian, Dong, Zhang, Zang, Ding, Lin, and Wang]{qian2024videostreaming}
Rui Qian, Xiaoyi Dong, Pan Zhang, Yuhang Zang, Shuangrui Ding, Dahua Lin, and Jiaqi Wang.
\newblock Streaming long video understanding with large language models.
\newblock \emph{NeurIPS}, 37:\penalty0 119336--119360, 2025.

\bibitem[Ren et~al.(2024)Ren, Yao, Li, Sun, and Hou]{ren2024timechat}
Shuhuai Ren, Linli Yao, Shicheng Li, Xu Sun, and Lu Hou.
\newblock Timechat: A time-sensitive multimodal large language model for long video understanding.
\newblock In \emph{CVPR}, pages 14313--14323, 2024.

\bibitem[Sermanet et~al.(2024)Sermanet, Ding, Zhao, Xia, Dwibedi, Gopalakrishnan, Chan, Dulac-Arnold, Maddineni, Joshi, et~al.]{sermanet2023robovqa}
Pierre Sermanet, Tianli Ding, Jeffrey Zhao, Fei Xia, Debidatta Dwibedi, Keerthana Gopalakrishnan, Christine Chan, Gabriel Dulac-Arnold, Sharath Maddineni, Nikhil~J Joshi, et~al.
\newblock Robovqa: Multimodal long-horizon reasoning for robotics.
\newblock In \emph{ICRA}, pages 645--652. IEEE, 2024.

\bibitem[Song et~al.(2024)Song, Chai, Wang, Zhang, Zhou, Wu, Chi, Guo, Ye, Zhang, et~al.]{moviechat}
Enxin Song, Wenhao Chai, Guanhong Wang, Yucheng Zhang, Haoyang Zhou, Feiyang Wu, Haozhe Chi, Xun Guo, Tian Ye, Yanting Zhang, et~al.
\newblock Moviechat: From dense token to sparse memory for long video understanding.
\newblock In \emph{CVPR}, pages 18221--18232, 2024.

\bibitem[Su et~al.(2024)Su, Ahmed, Lu, Pan, Bo, and Liu]{su2024rope}
Jianlin Su, Murtadha Ahmed, Yu Lu, Shengfeng Pan, Wen Bo, and Yunfeng Liu.
\newblock Roformer: Enhanced transformer with rotary position embedding.
\newblock \emph{Neurocomputing}, 568:\penalty0 127063, 2024.

\bibitem[Supancic~III and Ramanan(2017)]{supancic2017tracking}
James Supancic~III and Deva Ramanan.
\newblock Tracking as online decision-making: Learning a policy from streaming videos with reinforcement learning.
\newblock In \emph{ICCV}, pages 322--331, 2017.

\bibitem[Tan et~al.(2021)Tan, Zhang, Liu, Huang, Yang, Zhou, and Hu]{tan2021dynamic}
Qiaoyu Tan, Jianwei Zhang, Ninghao Liu, Xiao Huang, Hongxia Yang, Jingren Zhou, and Xia Hu.
\newblock Dynamic memory based attention network for sequential recommendation.
\newblock In \emph{AAAI}, pages 4384--4392, 2021.

\bibitem[Team et~al.(2024)Team, Georgiev, Lei, Burnell, Bai, Gulati, Tanzer, Vincent, Pan, Wang, et~al.]{team2024gemini}
Gemini Team, Petko Georgiev, Ving~Ian Lei, Ryan Burnell, Libin Bai, Anmol Gulati, Garrett Tanzer, Damien Vincent, Zhufeng Pan, Shibo Wang, et~al.
\newblock Gemini 1.5: Unlocking multimodal understanding across millions of tokens of context.
\newblock \emph{arXiv preprint arXiv:2403.05530}, 2024.

\bibitem[Team et~al.(2025)Team, Du, Yin, Xing, Qu, Wang, Chen, Zhang, Du, Wei, et~al.]{team2025kimivl}
Kimi Team, Angang Du, Bohong Yin, Bowei Xing, Bowen Qu, Bowen Wang, Cheng Chen, Chenlin Zhang, Chenzhuang Du, Chu Wei, et~al.
\newblock Kimi-vl technical report.
\newblock \emph{arXiv preprint arXiv:2504.07491}, 2025.

\bibitem[Touvron et~al.(2023{\natexlab{a}})Touvron, Lavril, Izacard, Martinet, Lachaux, Lacroix, Rozi{\`e}re, Goyal, Hambro, Azhar, Rodriguez, Joulin, Grave, and Lample]{touvron2023llama}
Hugo Touvron, Thibaut Lavril, Gautier Izacard, Xavier Martinet, Marie-Anne Lachaux, Timoth{\'e}e Lacroix, Baptiste Rozi{\`e}re, Naman Goyal, Eric Hambro, Faisal Azhar, Aurelien Rodriguez, Armand Joulin, Edouard Grave, and Guillaume Lample.
\newblock Llama: Open and efficient foundation language models.
\newblock \emph{arXiv preprint arXiv:2302.13971}, 2023{\natexlab{a}}.

\bibitem[Touvron et~al.(2023{\natexlab{b}})Touvron, Martin, Stone, Albert, Almahairi, Babaei, Bashlykov, Batra, Bhargava, Bhosale, et~al.]{Touvron2023Llama2O}
Hugo Touvron, Louis Martin, Kevin Stone, Peter Albert, Amjad Almahairi, Yasmine Babaei, Nikolay Bashlykov, Soumya Batra, Prajjwal Bhargava, Shruti Bhosale, et~al.
\newblock Llama 2: Open foundation and fine-tuned chat models.
\newblock \emph{arXiv preprint arXiv:2307.09288}, 2023{\natexlab{b}}.

\bibitem[Wang et~al.(2024{\natexlab{a}})Wang, Bai, Tan, Wang, Fan, Bai, Chen, Liu, Wang, Ge, et~al.]{wang2024qwen2vl}
Peng Wang, Shuai Bai, Sinan Tan, Shijie Wang, Zhihao Fan, Jinze Bai, Keqin Chen, Xuejing Liu, Jialin Wang, Wenbin Ge, et~al.
\newblock Qwen2-vl: Enhancing vision-language model's perception of the world at any resolution.
\newblock \emph{arXiv preprint arXiv:2409.12191}, 2024{\natexlab{a}}.

\bibitem[Wang et~al.(2024{\natexlab{b}})Wang, He, Hong, Cheng, Zhang, Qi, Gu, Huang, Xu, Dong, et~al.]{wang2024lvbench}
Weihan Wang, Zehai He, Wenyi Hong, Yean Cheng, Xiaohan Zhang, Ji Qi, Xiaotao Gu, Shiyu Huang, Bin Xu, Yuxiao Dong, et~al.
\newblock Lvbench: An extreme long video understanding benchmark.
\newblock \emph{arXiv preprint arXiv:2406.08035}, 2024{\natexlab{b}}.

\bibitem[Wang et~al.(2024{\natexlab{c}})Wang, Si, Wu, Zhu, Cao, and Nie]{wang2024retake}
Xiao Wang, Qingyi Si, Jianlong Wu, Shiyu Zhu, Li Cao, and Liqiang Nie.
\newblock Retake: Reducing temporal and knowledge redundancy for long video understanding.
\newblock \emph{arXiv preprint arXiv:2412.20504}, 2024{\natexlab{c}}.

\bibitem[Wang et~al.(2021)Wang, Chen, Jiang, Song, Han, and Huang]{wang2021adaptive}
Yulin Wang, Zhaoxi Chen, Haojun Jiang, Shiji Song, Yizeng Han, and Gao Huang.
\newblock Adaptive focus for efficient video recognition.
\newblock In \emph{ICCV}, pages 16249--16258, 2021.

\bibitem[Wang et~al.(2022{\natexlab{a}})Wang, Yue, Lin, Jiang, Lai, Kulikov, Orlov, Shi, and Huang]{wang2022adafocus}
Yulin Wang, Yang Yue, Yuanze Lin, Haojun Jiang, Zihang Lai, Victor Kulikov, Nikita Orlov, Humphrey Shi, and Gao Huang.
\newblock Adafocus v2: End-to-end training of spatial dynamic networks for video recognition.
\newblock In \emph{CVPR}, pages 20030--20040. IEEE, 2022{\natexlab{a}}.

\bibitem[Wang et~al.(2022{\natexlab{b}})Wang, Yue, Xu, Hassani, Kulikov, Orlov, Song, Shi, and Huang]{wang2022adafocusv3}
Yulin Wang, Yang Yue, Xinhong Xu, Ali Hassani, Victor Kulikov, Nikita Orlov, Shiji Song, Humphrey Shi, and Gao Huang.
\newblock Adafocusv3: On unified spatial-temporal dynamic video recognition.
\newblock In \emph{ECCV}, pages 226--243. Springer, 2022{\natexlab{b}}.

\bibitem[Wang et~al.(2024{\natexlab{d}})Wang, Zhang, Tang, Liu, Feng, Dai, and Jin]{wang2024hierar}
Yiqin Wang, Haoji Zhang, Yansong Tang, Yong Liu, Jiashi Feng, Jifeng Dai, and Xiaojie Jin.
\newblock Hierarchical memory for long video qa.
\newblock \emph{arXiv preprint arXiv:2407.00603}, 2024{\natexlab{d}}.

\bibitem[Wang et~al.(2024{\natexlab{e}})Wang, Zhang, Tian, and Tang]{wang2024ponder}
Yiqin Wang, Haoji Zhang, Jingqi Tian, and Yansong Tang.
\newblock Ponder \& press: Advancing visual gui agent towards general computer control.
\newblock \emph{arXiv preprint arXiv:2412.01268}, 2024{\natexlab{e}}.

\bibitem[Wang et~al.(2024{\natexlab{f}})Wang, Zhang, Yue, Song, Deng, Feng, and Huang]{wang2024uniada}
Yulin Wang, Haoji Zhang, Yang Yue, Shiji Song, Chao Deng, Junlan Feng, and Gao Huang.
\newblock Uni-adafocus: Spatial-temporal dynamic computation for video recognition.
\newblock \emph{TPAMI}, 2024{\natexlab{f}}.

\bibitem[Wang et~al.(2025{\natexlab{a}})Wang, Ni, Liu, Yuan, and Tang]{wang2025iterprime}
Yuji Wang, Jingchen Ni, Yong Liu, Chun Yuan, and Yansong Tang.
\newblock Iterprime: Zero-shot referring image segmentation with iterative grad-cam refinement and primary word emphasis.
\newblock In \emph{AAAI}, pages 8159--8168, 2025{\natexlab{a}}.

\bibitem[Wang et~al.(2025{\natexlab{b}})Wang, Xu, Liu, Li, and Tang]{wang2025sam2love}
Yuji Wang, Haoran Xu, Yong Liu, Jiaze Li, and Yansong Tang.
\newblock Sam2-love: Segment anything model 2 in language-aided audio-visual scenes.
\newblock In \emph{CVPR}, pages 28932--28941, 2025{\natexlab{b}}.

\bibitem[Wang et~al.(2020)Wang, Zheng, Liu, Li, and Wang]{track1}
Zhongdao Wang, Liang Zheng, Yixuan Liu, Yali Li, and Shengjin Wang.
\newblock Towards real-time multi-object tracking.
\newblock In \emph{ECCV}, pages 107--122, 2020.

\bibitem[Wu et~al.(2024)Wu, Chen, Lin, Wang, Gao, Xu, Xu, Hu, Chen, and Shou]{wu2024videollmmod}
Shiwei Wu, Joya Chen, Kevin~Qinghong Lin, Qimeng Wang, Yan Gao, Qianli Xu, Tong Xu, Yao Hu, Enhong Chen, and Mike~Zheng Shou.
\newblock Videollm-mod: Efficient video-language streaming with mixture-of-depths vision computation.
\newblock \emph{NeurIPS}, 37:\penalty0 109922--109947, 2024.

\bibitem[Xiao et~al.(2021)Xiao, Shang, Yao, and Chua]{xiao2021next}
Junbin Xiao, Xindi Shang, Angela Yao, and Tat-Seng Chua.
\newblock Next-qa: Next phase of question-answering to explaining temporal actions.
\newblock In \emph{CVPR}, pages 9777--9786, 2021.

\bibitem[Xue et~al.(2024)Xue, Chen, Li, Hu, Zhu, Li, Fang, Tang, Yang, Liu, et~al.]{xue2024longvila}
Fuzhao Xue, Yukang Chen, Dacheng Li, Qinghao Hu, Ligeng Zhu, Xiuyu Li, Yunhao Fang, Haotian Tang, Shang Yang, Zhijian Liu, et~al.
\newblock Longvila: Scaling long-context visual language models for long videos.
\newblock \emph{arXiv preprint arXiv:2408.10188}, 2024.

\bibitem[Yan et~al.(2019)Yan, Zhuang, Ni, Zhang, Xu, Zhang, Zhang, Cheng, Tian, Xu, et~al.]{yan2019finegrain}
Yichao Yan, Ning Zhuang, Bingbing Ni, Jian Zhang, Minghao Xu, Qiang Zhang, Zheng Zhang, Shuo Cheng, Qi Tian, Yi Xu, et~al.
\newblock Fine-grained video captioning via graph-based multi-granularity interaction learning.
\newblock \emph{TPAMI}, 44\penalty0 (2):\penalty0 666--683, 2019.

\bibitem[Yang et~al.(2024{\natexlab{a}})Yang, Yang, Hui, Zheng, Yu, Zhou, Li, Li, Liu, Huang, et~al.]{yang2024qwen2}
An Yang, Baosong Yang, Binyuan Hui, Bo Zheng, Bowen Yu, Chang Zhou, Chengpeng Li, Chengyuan Li, Dayiheng Liu, Fei Huang, et~al.
\newblock Qwen2 technical report.
\newblock \emph{arXiv preprint arXiv:2407.10671}, 2024{\natexlab{a}}.

\bibitem[Yang et~al.(2024{\natexlab{b}})Yang, Wang, Ye, Tang, Chen, Zhao, and Torr]{yang2024lavt}
Zhao Yang, Jiaqi Wang, Xubing Ye, Yansong Tang, Kai Chen, Hengshuang Zhao, and Philip~HS Torr.
\newblock Language-aware vision transformer for referring segmentation.
\newblock \emph{TPAMI}, 2024{\natexlab{b}}.

\bibitem[Ye et~al.(2025{\natexlab{a}})Ye, Gan, Ge, Zhang, and Tang]{ye2025atp}
Xubing Ye, Yukang Gan, Yixiao Ge, Xiao-Ping Zhang, and Yansong Tang.
\newblock Atp-llava: Adaptive token pruning for large vision language models.
\newblock In \emph{CVPR}, pages 24972--24982, 2025{\natexlab{a}}.

\bibitem[Ye et~al.(2025{\natexlab{b}})Ye, Gan, Huang, Ge, and Tang]{ye2025voco}
Xubing Ye, Yukang Gan, Xiaoke Huang, Yixiao Ge, and Yansong Tang.
\newblock Voco-llama: Towards vision compression with large language models.
\newblock In \emph{CVPR}, pages 29836--29846, 2025{\natexlab{b}}.

\bibitem[Yu et~al.(2023)Yu, Cho, Yadav, and Bansal]{yu2024sevila}
Shoubin Yu, Jaemin Cho, Prateek Yadav, and Mohit Bansal.
\newblock Self-chained image-language model for video localization and question answering.
\newblock \emph{Advances in Neural Information Processing Systems}, 36:\penalty0 76749--76771, 2023.

\bibitem[Yu et~al.(2019)Yu, Xu, Yu, Yu, Zhao, Zhuang, and Tao]{yu2019activitynet}
Zhou Yu, Dejing Xu, Jun Yu, Ting Yu, Zhou Zhao, Yueting Zhuang, and Dacheng Tao.
\newblock Activitynet-qa: A dataset for understanding complex web videos via question answering.
\newblock In \emph{AAAI}, pages 9127--9134, 2019.

\bibitem[Zhang et~al.(2016)Zhang, Wang, Wang, Qiao, and Wang]{action1}
Bowen Zhang, Limin Wang, Zhe Wang, Yu Qiao, and Hanli Wang.
\newblock Real-time action recognition with enhanced motion vector cnns.
\newblock In \emph{CVPR}, pages 2718--2726, 2016.

\bibitem[Zhang et~al.(2024{\natexlab{a}})Zhang, Zhang, Li, Zeng, Yang, Zhang, Wang, Tan, Li, and Liu]{zhang2024longva}
Peiyuan Zhang, Kaichen Zhang, Bo Li, Guangtao Zeng, Jingkang Yang, Yuanhan Zhang, Ziyue Wang, Haoran Tan, Chunyuan Li, and Ziwei Liu.
\newblock Long context transfer from language to vision.
\newblock \emph{arXiv preprint arXiv:2406.16852}, 2024{\natexlab{a}}.

\bibitem[Zhang et~al.(2024{\natexlab{b}})Zhang, Wu, Li, Li, Ma, Liu, and Li]{zhang2024llavavideo}
Yuanhan Zhang, Jinming Wu, Wei Li, Bo Li, Zejun Ma, Ziwei Liu, and Chunyuan Li.
\newblock Video instruction tuning with synthetic data.
\newblock \emph{arXiv preprint arXiv:2410.02713}, 2024{\natexlab{b}}.

\bibitem[Zhou et~al.(2024{\natexlab{a}})Zhou, Shu, Zhao, Wu, Xiao, Yang, Xiong, Zhang, Huang, and Liu]{zhou2024mlvu}
Junjie Zhou, Yan Shu, Bo Zhao, Boya Wu, Shitao Xiao, Xi Yang, Yongping Xiong, Bo Zhang, Tiejun Huang, and Zheng Liu.
\newblock Mlvu: A comprehensive benchmark for multi-task long video understanding.
\newblock \emph{arXiv preprint arXiv:2406.04264}, 2024{\natexlab{a}}.

\bibitem[Zhou et~al.(2024{\natexlab{b}})Zhou, Arnab, Buch, Yan, Myers, Xiong, Nagrani, and Schmid]{zhou2024streamingcaption}
Xingyi Zhou, Anurag Arnab, Shyamal Buch, Shen Yan, Austin Myers, Xuehan Xiong, Arsha Nagrani, and Cordelia Schmid.
\newblock Streaming dense video captioning.
\newblock In \emph{CVPR}, pages 18243--18252, 2024{\natexlab{b}}.

\bibitem[Zhu et~al.(2025)Zhu, Wang, Li, Zhao, Tang, Niu, Chen, Zhou, and Lu]{zhu2025instarevive}
Yixuan Zhu, Haolin Wang, Ao Li, Wenliang Zhao, Yansong Tang, Jingxuan Niu, Lei Chen, Jie Zhou, and Jiwen Lu.
\newblock Instarevive: One-step image enhancement via dynamic score matching.
\newblock \emph{arXiv preprint arXiv:2504.15513}, 2025.

\end{thebibliography}
    }
    
    % WARNING: do not forget to delete the supplementary pages from your submission 
    \clearpage
\maketitlesupplementary
\appendix

\noindent
In the supplementary material, we first provide implementation details of the Flash Memory mechanism and training  settings. 
Subsequently, we conduct an analysis experiment on model inference efficiency and more ablation studies on memory structure configurations.
We then present more visual cases to provide a comprehensive understanding of the performance of models.
% We highly recommend watching \textbf{the supplementary video}, which contains a live demonstration of real-time multimodal assistant based on Flash-VStream model.

\section{Implementation Details}

This section describes the details of the proposed Flash Memory mechanism in Sec. 3. The Flash Memory consists of Context Synopsis Memory (CSM) and Detail Augmentation Memory (DAM). CSM uses a clustering-based updating policy, while DAM uses a retrieval-based updating policy.

\begin{align}
M^{\text{CSM}}_k &= \frac{1}{|S_k|} \sum_{i \in S_k} e_i^{\text{L}}, 1 \leq k \leq N^\text{CSM} \\
M^{\text{CSM}} &=\operatorname{cluster} (M^{\text{CSM}} \oplus e_{t+1}^{\text{L}})\label{eq:supp_cluster}
\end{align}

\subsection{Context Synopsis Memory}
As mentioned in Sec. 3.2, CSM is designed for aggregating long-context temporal information and modeling the distribution of information density.
$M^{\text{CSM}}_k$ represents the centroid of the k-th cluster. $M^{\text{CSM}}$ is initialized with the first $N^{\text{CSM}}$ feature maps of the first $N^{\text{CSM}}$ frames. When the next frame arrives, a clustering algorithm is employed to consolidate its feature map into existing clusters. Here we illustrate the ``cluster'' operation of \Cref{eq:supp_cluster} in detail.

As shown in \Cref{alg:wkmeans}, CSM performs a temporal-wise \textit{K-means Clustering} algorithm to condense $(N^{\text{CSM}}+1) \times h^\prime \times w^\prime$ tokens to $N^{\text{CSM}} \times h^\prime \times w^\prime$ tokens. Each frame feature in temporal memory 
$M_{k}^{\text{CSM}} = c_{k} \in \mathbb{R}^{h^\prime \times w^\prime \times d}$ 
represents the centroid of the i-th feature map cluster.

\begin{algorithm}[t]
    \footnotesize    
    \caption{K-means Clustering Algorithm}
    \label{alg:wkmeans}
    \begin{algorithmic}[1] % The number tells where the line numbering should start
        \Require Current cluster centroids $M=M^{\text{CSM}}$
        \Require Newest frame feature $e=e_t^{\text{L}}$
        \Require Set of all points $X = \{ M_1, M_2, \dots, M_{N}, e \}$
        \Require Weights vector of points $W = \{w_1, w_2, \dots, w_{N}, 1\}$
        \Require Maximum memory length $N=N^{\text{CSM}}$
        \Require Maximum number of iterations $T$
        \Procedure{K-means}{$X, W, N, T$}
            \State Initialize $t \gets 0$
            \State Initialize centroids $C=\{c_1, c_2, \dots, c_N\}$ from $X$
            \State Initialize cluster assignment $S_j \gets \{ \}, 1 \leq j \leq N$
            % \State Initialize previous assignment $P_j \gets \{ \}, 1 \leq j \leq N$
            \While{$t < T$}
                \For{ $x_i \in X$}
                    \State $j \gets \underset{j}{\operatorname{argmin}} \lVert x_i - c_j {\rVert}^2$
                    \State $S_j \gets S_j \cup \{x_i\}$
                \EndFor
                % \If{$S == P$}
                %     \State \textbf{break}
                % \EndIf
                \For{ $j = 1, 2, \dots, N$}
                    \State $ \displaystyle c_j^{\text{new}} \gets \frac{\sum_{x_i \in S_j} w_i \cdot x_i}{\sum_{x_i \in S_j} w_i}$
                \EndFor
                % \State $P \gets S$
                \State Clear $S$
                \State $C \gets C^{\text{new}}$
                \State $t \gets t + 1$
            \EndWhile 
            \For{ $j = 1, 2, \dots, N$}
                \State $ w_j^{\text{CSM}} \gets \sum_{x_i \in S_j} w_i$
            \EndFor
            \State $M^{\text{CSM}} = C$
            \State $W^{\text{CSM}} = \{w_1^{\text{CSM}}, w_2^{\text{CSM}}, \dots, w_N^{\text{CSM}}\}$
            \State \Return $M^{\text{CSM}}, W^{\text{CSM}}$
        \EndProcedure
    \end{algorithmic}
\end{algorithm}

\begin{algorithm}[t]
    \footnotesize
    \caption{Feature-Centric Sampling}
    \label{alg:ret}
    \begin{algorithmic}[1] 
        \Require Current feature bank $E^{\text{H}}_t = \{e^{\text{H}}_1, e^{\text{H}}_2, \dots, e^{\text{H}}_t \} $
        \Require Current cluster centroids $M^{\text{CSM}}$
        \Require Weights vector of points $W = \{w_1, w_2, \dots, w_{N}\}$
        \Require Maximum memory length $N=N^{\text{DAM}}$
        \Procedure{Key Feature Retrieval}{$E^{\text{H}}, M^{\text{CSM}}, W, N$}       % Python-style function
            \State $k \gets  N$
            \State $idx \gets \operatorname{argsort}(W, \text{descending=True})$  % 第一步：计算argsort
            \State $j_1, j_2, \dots, j_k \gets idx[:k]$  % 第二步：选择前k个
    
            \State $M^{\text{DAM}} \gets \{ \}$
            \For{$z = 1, 2, \dots, k$}
                \State $anchor \gets M^{\text{CSM}}_{j_z}$
                \State $\displaystyle i \gets \underset{i \leq t}{\operatorname{argmin}} \; \lVert e^{\text{L}}_i - anchor \rVert^2$
                \State $M^{\text{DAM}} \gets M^{\text{DAM}} \cup \{e^{\text{H}}_i\}$
            \EndFor
            \State \Return $M^{\text{DAM}}$
        \EndProcedure
    \end{algorithmic}
\end{algorithm}

\subsection{Detail Augmentation Memory}
As described in Sec. 3.3, DAM aims at storing spatial details of the most informative key frames, based on the feature clusters of CSM.
For DAM, we use a Feature-Centric Sampling method to calculate $M^{\text{DAM}} \in \mathbb{R}^{N^{\text{DAM}} \times h \times w \times d}$. 

\Cref{alg:ret} shows the pseudo code of \textit{Feature-Centric Retrieval}.
Here $w_j$ is equal to the size of $j$-th cluster, i.e., the number of feature maps in this cluster. 
We choose the centroids of the top-k largest clusters as anchors. Then we select key features from the feature bank $E^{H}_t$. 
$E^{H}_t$ keeps high-resolution feature maps of all frames on disk, where $t$ is the current number of frames.
The features nearest to these anchors in the feature map space are considered as key features, which are added to the DAM.

\begin{figure*}[h]
    \centering
    \begin{subfigure}[b]{0.33\linewidth}
        \centering
        \includegraphics[width=\linewidth]{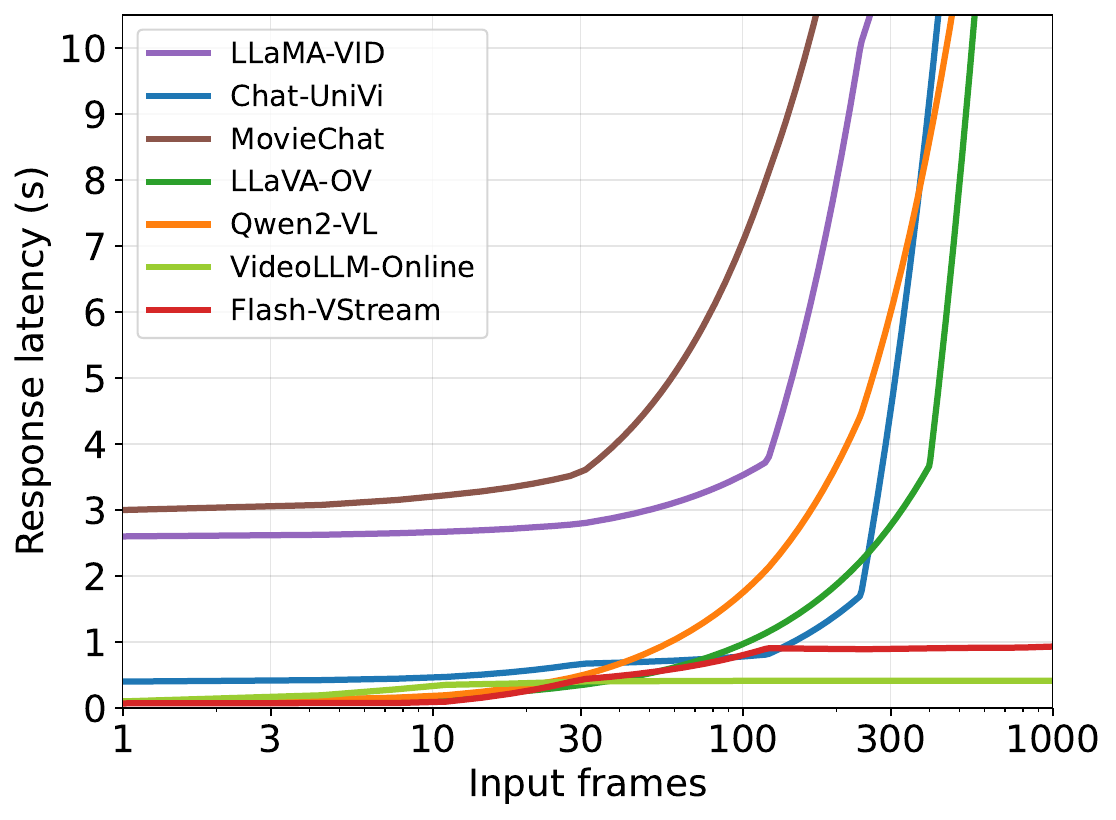}
        \caption{Response latency (s) v.s. Input frames}
        \label{fig:supp_efficiency_a}
    \end{subfigure}
    \hfill
    \begin{subfigure}[b]{0.33\linewidth}
        \centering
        \includegraphics[width=\linewidth]{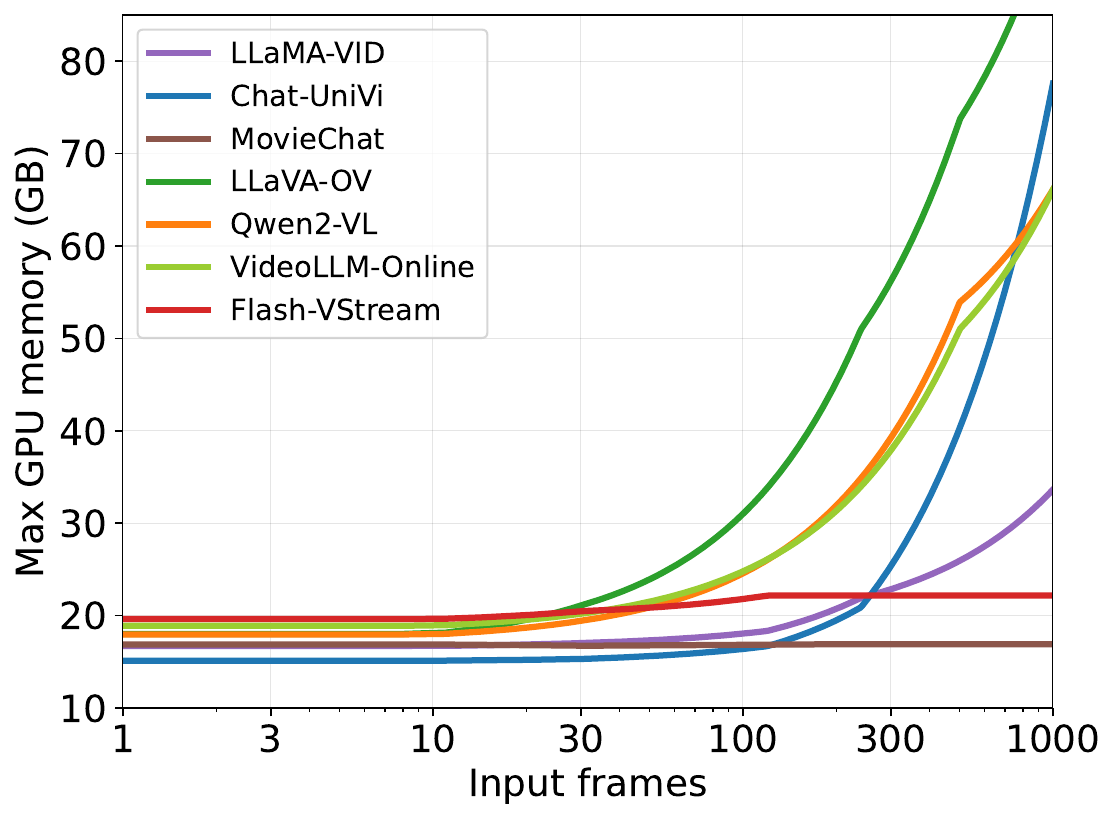}
        \caption{Max GPU memory (GB) v.s. Input frames}
        \label{fig:supp_efficiency_b}
    \end{subfigure}
    \hfill
    \begin{subfigure}[b]{0.325\linewidth}
        \centering
        \includegraphics[width=\linewidth]{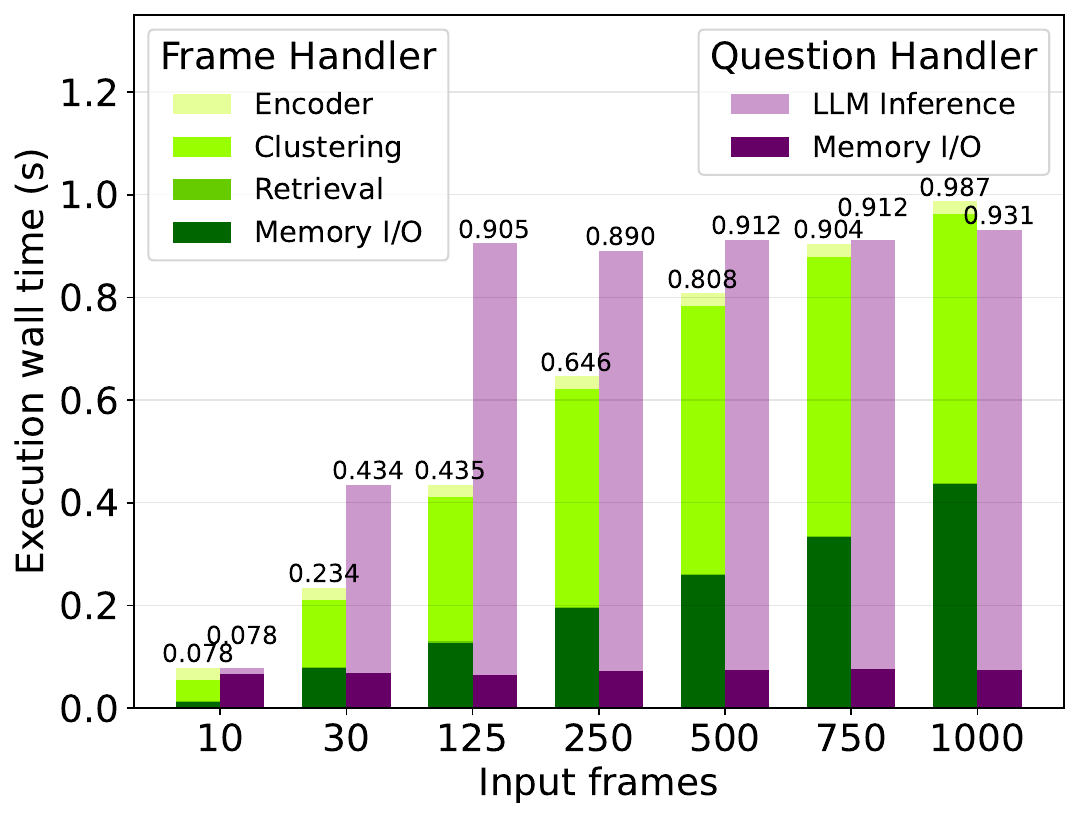}
        \caption{Execution wall time (s) v.s. Input frames}
        \label{fig:supp_efficiency_c}
    \end{subfigure}
    \caption{\textbf{(a) Response latency comparison. (b) Max GPU memory comparison. (c) Execution wall time analysis.}
    Response latency refers to the wall time between inputting a question and outputting the first token of the answer.
    Max GPU memory indicates the peak GPU memory usage during inference.
    All experiments were conducted on A100 GPUs using BFloat16 and FlashAttention-2.
    }
    \label{fig:supp_efficiency}
\end{figure*}

\begin{table}[h]
    \centering
    \begin{tabular}{ll}
        \Xhline{0.8px}
            \rowcolor{lightblue}
            \textbf{Settings}   & \textbf{Value} \\
        \hline
            Batch Size & 64 \\ 
            Learning Rate & 8e-4 \\ 
            Lora Rank & 64 \\ 
            Lora Alpha & 32 \\ 
            Learning Schedule & Cosine decay \\ 
            Warmup Ratio & 0.01 \\ 
            Weight Decay & 0.1 \\ 
            Epoch  & 1 \\ 
            Optimizer & AdamW \\ 
            Deepspeed Stage & 2 \\ 
            Visual Encoder & Freeze \\ 
            Projector & Open \\ 
            LLM & Open \\ 
        \Xhline{0.8px}
    \end{tabular}
    \caption{Training settings of Flash-VStream.}
    \label{tab:train_setting}
\end{table}

\section{Training Details}

We train Flash-VStream on a 9k subset of LLaVA-Video~\cite{zhang2024llavavideo} dataset for one epoch. During training, we freeze the parameters of visual encoder, while all linear layers of projector and LLM are LoRA finetuned.
The overall training can be finished in about 10 hours on 8 A100 80G GPUs with BFloat16 automatic mixed precision and FlashAttention-2~\cite{dao2024flashattention2}. Detailed training settings are shown in \Cref{tab:train_setting}.

\section{Efficiency Analysis}
\label{sec:C}
An efficiency analysis is performed to assess the inference efficiency of Flash-VStream. 
Specifically, we concentrate on the response latency and GPU memory consumption of models, as discussed in Sec. 1 of the paper.

We compare Flash-VStream with other competitive video language models~\cite{llamavid,chatunivi,moviechat,li2024llavaov,wang2024qwen2vl} in terms of response latency and max GPU memory.
As presented in~\Cref{fig:supp_efficiency}, Flash-VStream demonstrates superior performance in both efficiency metrics.
\Cref{fig:supp_efficiency_a} shows the response latency comparison, where Flash-VStream consistently exhibits lower latency across varying numbers of input frames. This indicates that Flash-VStream is more efficient in processing video inputs, resulting in faster response times (less than 1 second).
\Cref{fig:supp_efficiency_b} illustrates the maximum GPU memory usage. Flash-VStream maintains a relatively stable and lower GPU memory consumption compared to other models, even as the number of input frames increases. This efficiency in memory usage makes Flash-VStream more scalable and suitable for deployment in resource-constrained environments.

From a systematic perspective, we measure the execution wall time of each process in~\Cref{fig:supp_efficiency_c}. The result shows that the question handler process stays fast enough (< 1s) regardless of the number of input frames. This is because the question handler only relies on size-fixed Flash Memory.
The execution time of the frame handler process grows up to more than 1 second when the number of frames exceeds 1000. Although this may result in delayed updates of visual information, it would not affect the response latency.

Overall, the results highlight the efficiency advantages of Flash-VStream in terms of both response latency and GPU memory consumption, making it a competitive choice for real-time long video understanding tasks.

\section{Ablation Study on Memory Structure}

In Sec. 4.4 and Fig. 4, we initially explored the relationship between memory allocation strategy and pool ratio of CSM and DAM. Empirically, we found the best setting for these configurations under the fixed-budget constraint.
In this section, we aim to answer the following questions:

\noindent \textbf{Q1:} How sensitive is the model performance to cluster numbers of CSM, i.e., $N^{\text{CSM}}$?

\noindent \textbf{Q2:} How sensitive is the model performance to key frame numbers of DAM, i.e., $N^{\text{DAM}}$?

\begin{table*}[t]
    \centering
    \small
    \setlength{\tabcolsep}{8pt}
    \begin{tabular}{c|ccccc|cccc}
        % \toprule
        \Xhline{0.8px}
            \rowcolor{lightblue}
                & \multicolumn{5}{c|}{\textbf{Memory Component Settings}} & 
                \multicolumn{4}{c}{\textbf{Evaluation Results}} \\
            \rowcolor{lightblue}
                \multirow{-2}{*}{\textbf{ID}} & \textbf{CSM} & \textbf{DAM} & 
                \textbf{N\textsubscript{Vtokens}} & \textbf{CSM Size} & \textbf{DAM Size} & 
                \textbf{EgoSchema} & \textbf{MVBench} & \textbf{Video-MME(w/o)} & \textbf{Average} \\
        \hline
             & \ding{51} & \ding{51} & 19200 & $60 \times 64$ & $60 \times 256$ & 68.6 & 65.5 & 61.2 & 65.1 \\
             & \ding{51} & \ding{51} & 15360 & $60 \times 64$ & $45 \times 256$ & 68.3 & 65.3 & 61.0 & 64.9 \\
            \rowcolor{gray!10}
            \ding{172} & \ding{51} & \ding{51} & 11520 & $60 \times 64$ & $30 \times 256$ & 68.2 & 65.4 & 61.2 & 64.9 \\
             & \ding{51} & \ding{51} & 7680 & $60 \times 64$ & $15 \times 256$ & 67.5 & 64.9 & 60.8 & 64.4 \\
            \ding{174} & \ding{51} & \ding{55} & 3840  & $60 \times 64$ & $0$ & 66.8 & 64.0 & 60.1 & 63.6 \\
        \hline
             & \ding{51} & \ding{55} & 5760  & $90 \times 64$ & $0$ & 66.6 & 63.9 & 61.0 & 63.8 \\
            \rowcolor{gray!10}
            \ding{174} & \ding{51} & \ding{55} & 3840  & $60 \times 64$ & $0$ & 66.8 & 64.0 & 60.1 & 63.6 \\
             & \ding{51} & \ding{55} & 1920  & $30 \times 64$ & $0$ & 65.7 & 63.6 & 58.8 & 62.7 \\
             & \ding{51} & \ding{55} & 960  & $15 \times 64$ & $0$ & 63.0 & 63.0 & 58.3 & 61.5 \\
        % \bottomrule
        \Xhline{0.8px}
    \end{tabular}
    \caption{\textbf{Ablation study of memory structure configurations.} 
    We investigate the model's sensitivity to cluster numbers of CSM and key frame numbers of DAM.
    }
    \label{tab:supp_abl}
\end{table*}

\begin{table*}[t]
    \centering
    \small
    \setlength{\tabcolsep}{8pt}
    \begin{tabular}{c|ccccccc|c}  
        \Xhline{0.8px}
            \rowcolor{lightblue}
            \textbf{Score} & \textbf{0} & \textbf{1} & \textbf{2} & \textbf{3} & \textbf{4} & \textbf{5} & \textbf{Total} & \textbf{Average Score} \\ 
        \hline
            Right & 8 & 0 & 26 & 111 & 1916 & 2732 & 4793 & 4.53 \\ 
            Wrong & 355 & 290 & 1712 & 82 & 82 & 686 & 3207 & 2.41 \\ 
            Total & 363 & 290 & 1738 & 193 & 1998 & 3418 & 8000 & 3.68 \\ 
        \Xhline{0.8px}
    \end{tabular}
    \caption{\textbf{Score distribution of a GPT-3.5-based evaluation.}
    We tested Qwen2-VL-7b on ActivityNet-QA benchmark, using GPT-3.5-turbo-0125 for evaluation.
    It is observed that many wrong predictions are assigned with a high score ``5'', leading to a biased result.
    }
    \label{tab:supp_gpt35}
\end{table*}

As presented in~\Cref{tab:supp_abl}, we conduct two groups of experiments to investigate the model's sensitivity to memory structure configurations, i.e., memory sizes $N^{\text{CSM}}$ and $N^{\text{DAM}}$.
In each group, we compare different memory size choices to the baseline row \ding{172} and row \ding{174} in Table 4.
The results show a scaling trend of accuracy with different memory sizes.
Therefore, the results of grid search experiment illustrated in Fig. 4 are reasonable.

\section{Case Study}

In this section, we conduct a case study to provide a comprehensive understanding of the performance of models.
This study presents a series of visual cases involving various types of videos, each accompanied by a specific question and multiple-choice options to evaluate the performance of three different models: Qwen2-VL~\cite{wang2024qwen2vl}, LLaVA-OV~\cite{li2024llavaov}, and the proposed Flash-VStream. 

\begin{figure*}[t]
    \centering
    \includegraphics[width=\linewidth]{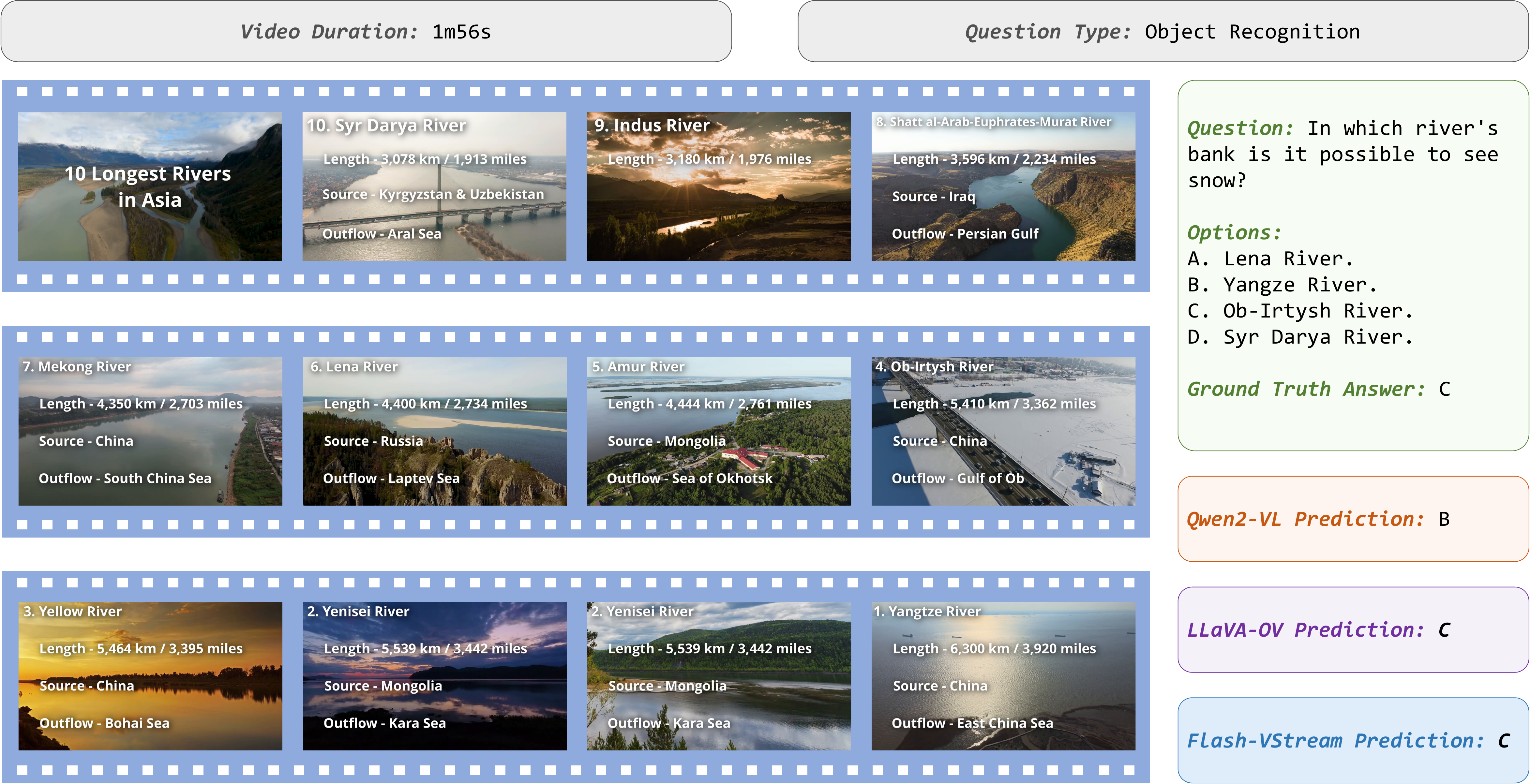}
    \caption{\textbf{Case Study.}
    This figure presents a case study on documentary video about the 10 longest rivers in Asia, highlighting their lengths, sources, and outflows. The study includes a question regarding the possibility of seeing snow on the banks of these rivers, with multiple-choice options provided. The ground truth answer is indicated, along with the predictions from three different models: Qwen2-VL, LLaVA-OV, and Flash-VStream.
    }
    \label{fig:supp_case_1}
\end{figure*}

\begin{figure*}[t]
    \centering
    \includegraphics[width=\linewidth]{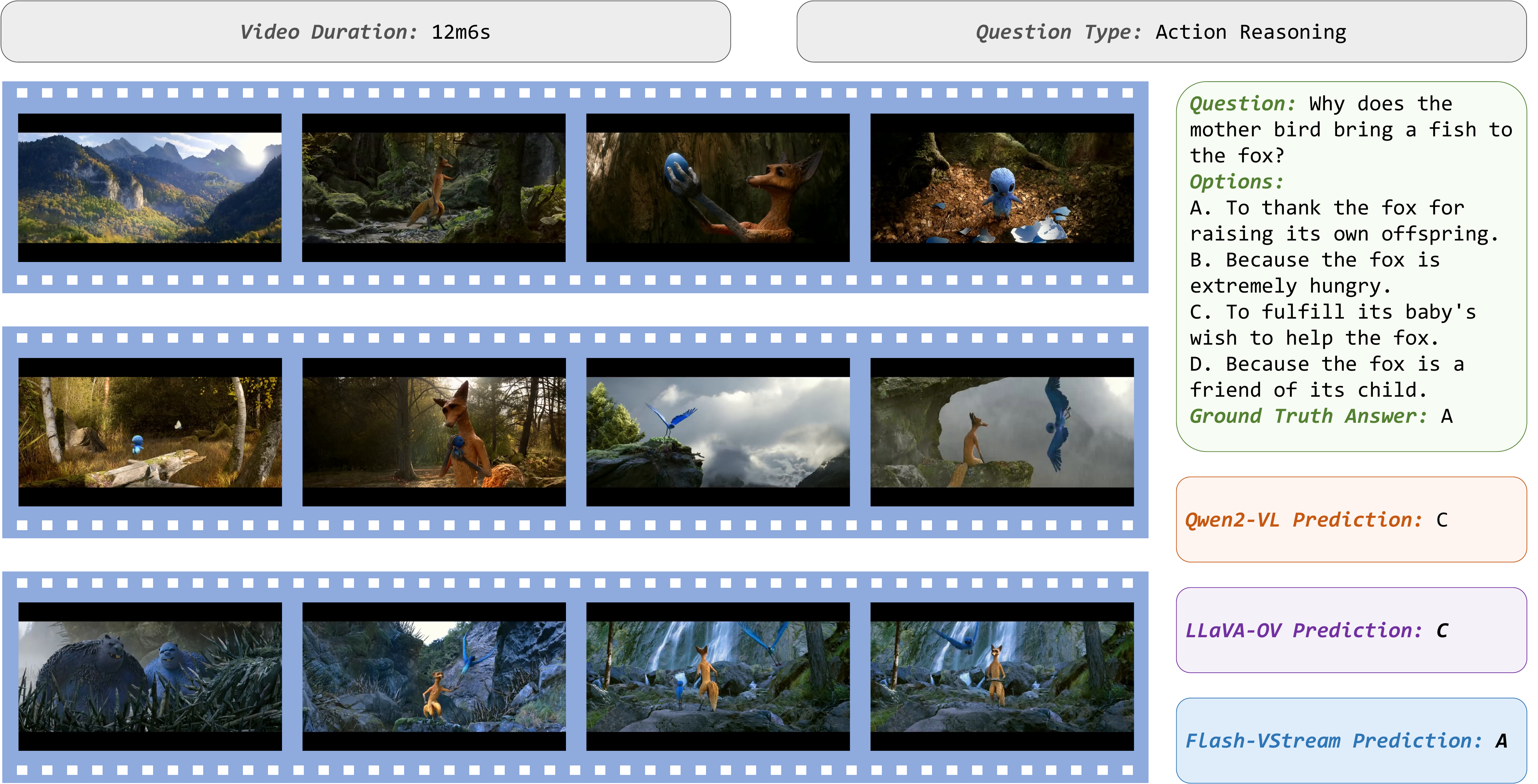}
    \caption{\textbf{Case Study.}
    This figure presents a case study involving a cartoon video depicting a mother bird bringing a fish to a fox. The study includes a question about the reason behind this action, with multiple-choice options provided. The ground truth answer is indicated, along with the predictions from three different models: Qwen2-VL, LLaVA-OV, and Flash-VStream.
    }
    \label{fig:supp_case_2}
\end{figure*}

\begin{figure*}[t]
    \centering
    \includegraphics[width=\linewidth]{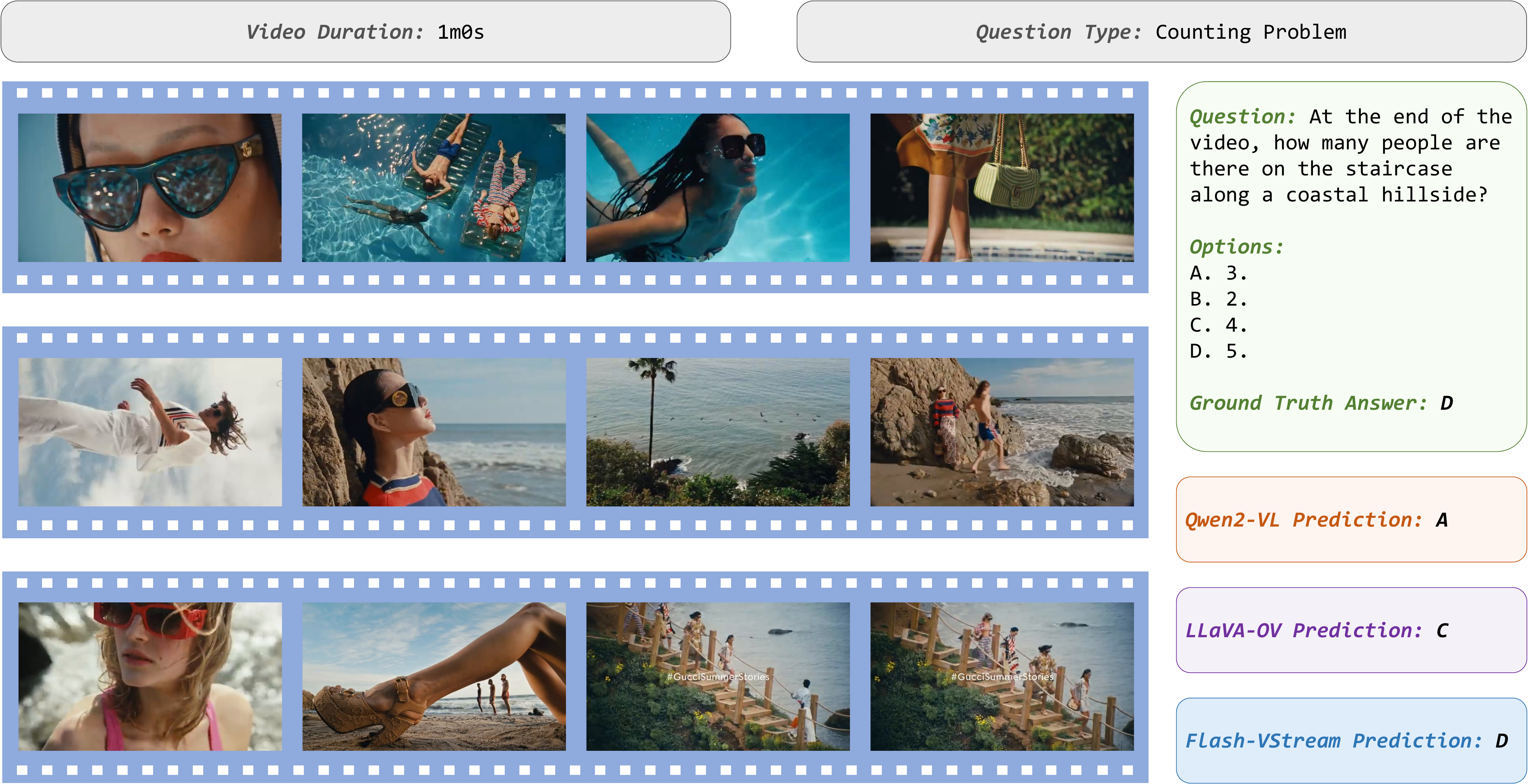}
    \caption{\textbf{Case Study.}
    This figure presents a case study involving an advertising video, depicting various scenes including people by the pool, on the beach, and along a coastal hillside. The study includes a question about the number of people on the staircase at the end of the video, with multiple-choice options provided. The ground truth answer is indicated, along with the predictions from three different models: Qwen2-VL, LLaVA-OV, and Flash-VStream.
    }
    \label{fig:supp_case_4}
\end{figure*}

\begin{figure*}[t]
    \centering
    \includegraphics[width=\linewidth]{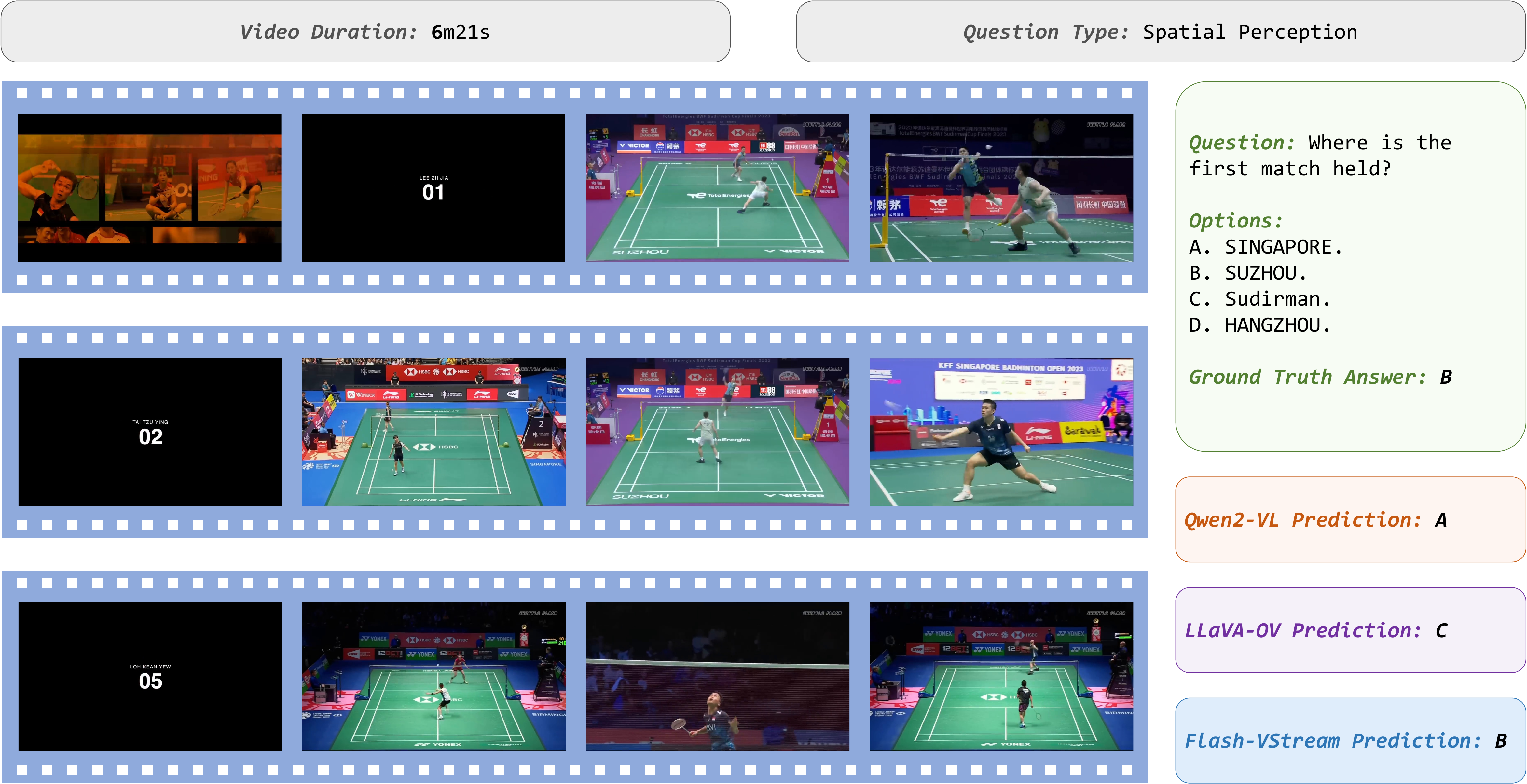}
    \caption{\textbf{Case Study.}
    This figure presents a case study involving a sports documentary video of badminton tournaments, depicting various matches and players. The study includes a question about the location of the first match, with multiple-choice options provided. The ground truth answer is indicated, along with the predictions from three different models: Qwen2-VL, LLaVA-OV, and Flash-VStream.
    }
    \label{fig:supp_case_5}
\end{figure*}

\begin{figure*}[t]
    \vspace{20pt}
    \centering
    \includegraphics[width=\linewidth]{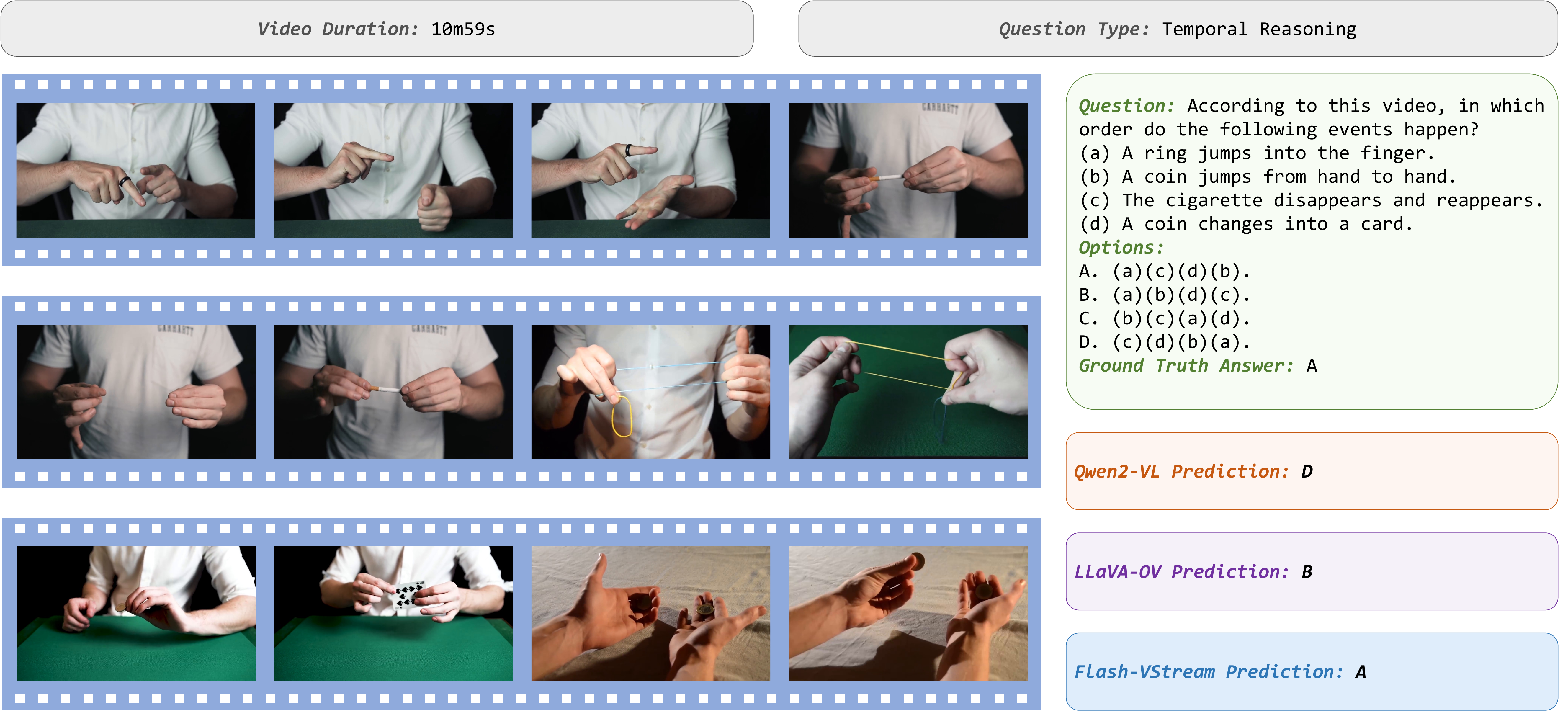}
    \caption{\textbf{Case Study.}
    This figure presents a case study involving a tutorial video depicting various magic tricks. The study includes a question about the order of events in the video, with multiple-choice options provided. The ground truth answer is indicated, along with the predictions from three different models: Qwen2-VL, LLaVA-OV, and Flash-VStream.
    }
    \label{fig:supp_case_3}
    \vspace{20pt}
\end{figure*}

\begin{figure*}[t]
    \centering
    \includegraphics[width=\linewidth]{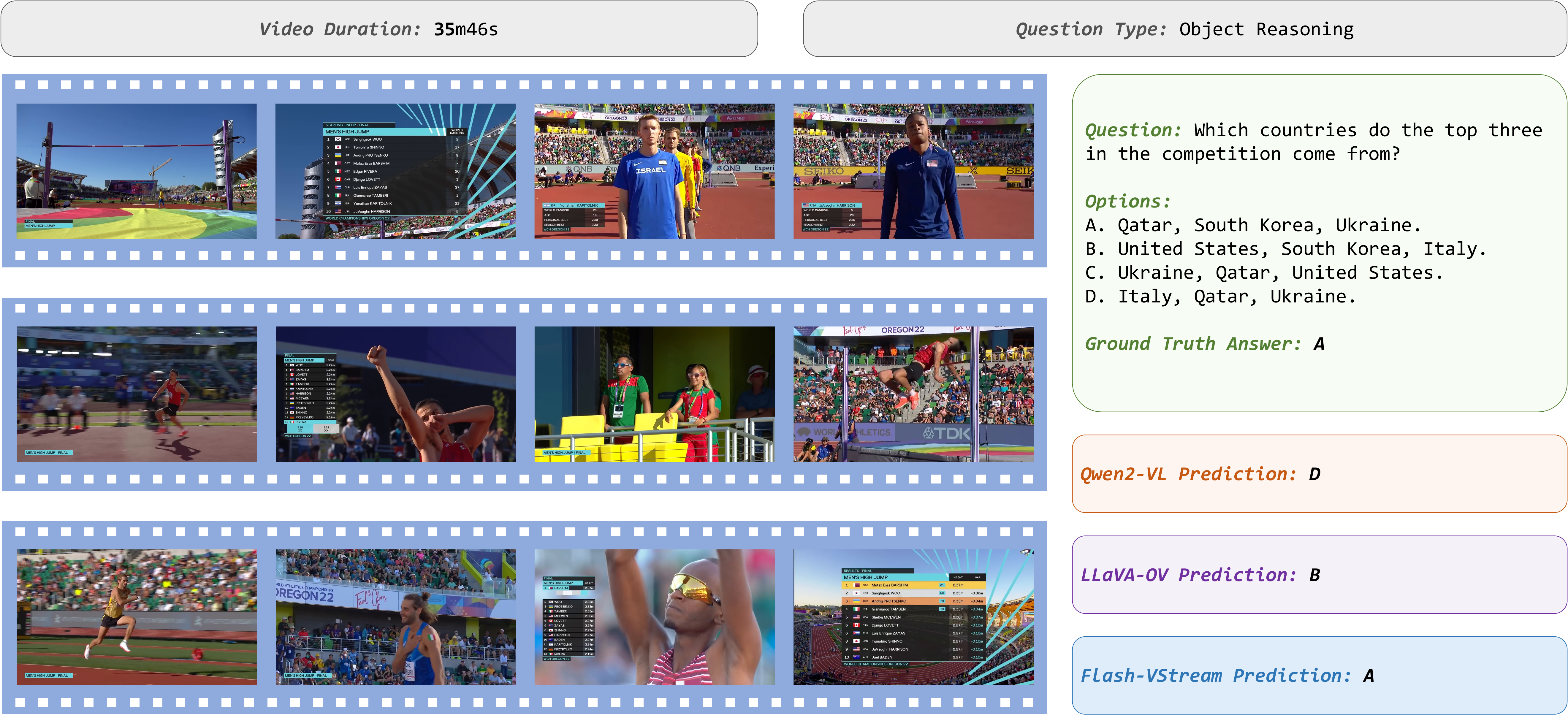}
    \caption{\textbf{Case Study.}
    This figure presents a case study involving a sports video from a high jump competition, depicting various athletes and their performances. 
    The video frames capture moments of intense competition, showcasing the athletes' skills and determination as they strive to achieve their best performances. The analysis aims to evaluate the models' ability to accurately interpret and predict the outcomes based on visual and contextual cues from the video.
    The study includes a question about the countries of the top three athletes in the competition, with multiple-choice options provided. The ground truth answer is indicated, along with the predictions from three different models: Qwen2-VL, LLaVA-OV, and Flash-VStream.
    }
    \label{fig:supp_case_6}
\end{figure*}

\Cref{fig:supp_case_1,fig:supp_case_2,fig:supp_case_3,fig:supp_case_4,fig:supp_case_5,fig:supp_case_6} present different genres of videos, including documentaries, cartoons, commercials, sports programs and tutorial videos. As shown in these cases, Flash-VStream exhibits strong understanding capabilities in object recognition, action recognition, action reasoning, temporal reasoning, object counting and object reasoning.




\begin{figure*}[t]
    \vspace{20pt}
    \centering
    \includegraphics[width=\linewidth]{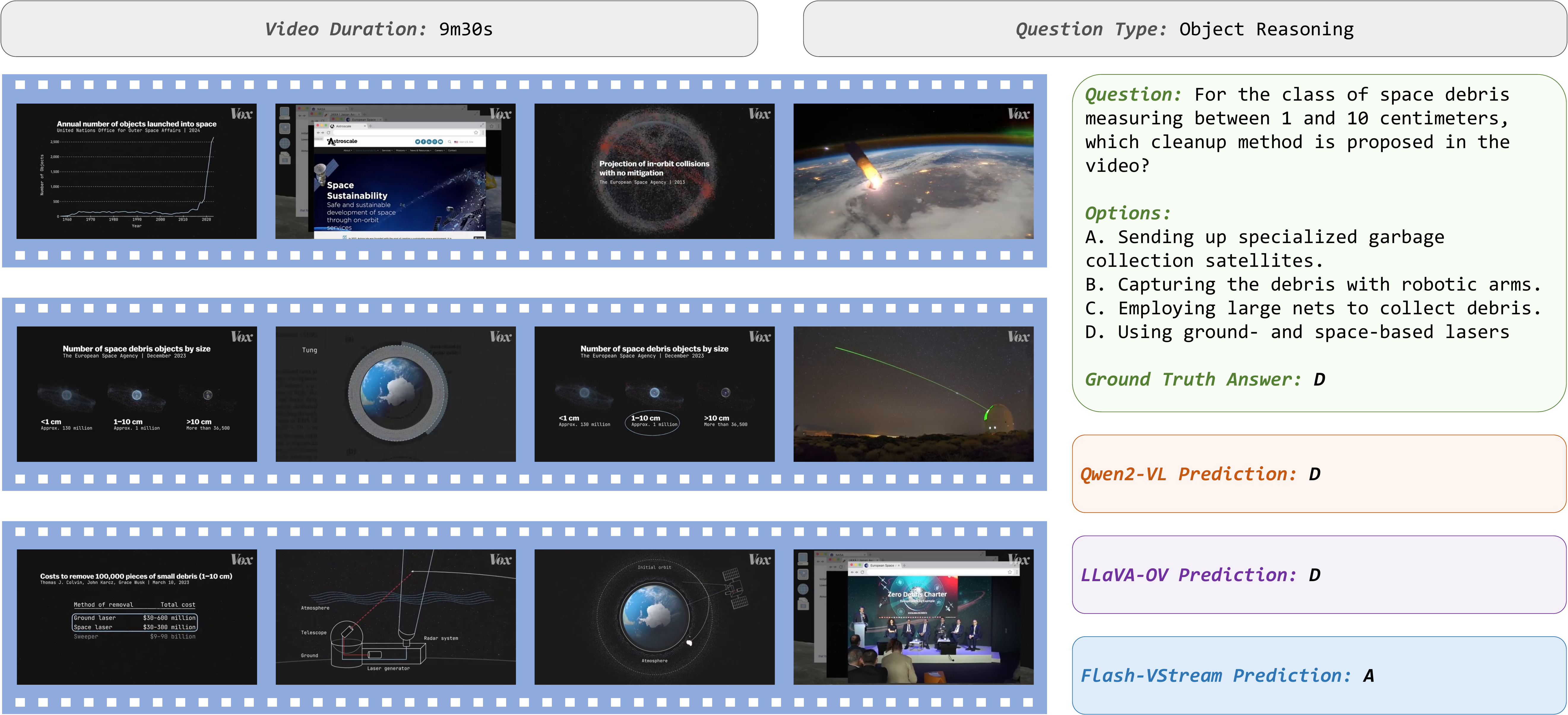}
    \caption{\textbf{Fail Case Analysis.}
    This figure presents a case study involving a video on space debris and proposed cleanup methods.  The video frames illustrate various statistics and methods related to space debris, highlighting the challenges and potential solutions for mitigating the growing problem of space junk. The study includes a question about the recommended method for cleaning up space debris measuring between 1 and 10 centimeters, with multiple-choice options provided. The ground truth answer is indicated, along with the predictions from three different models: Qwen2-VL, LLaVA-OV, and Flash-VStream.
    }
    \vspace{20pt}
    \label{fig:supp_fail_case_1}
\end{figure*}
\begin{figure*}[t]
    \centering
    \includegraphics[width=\linewidth]{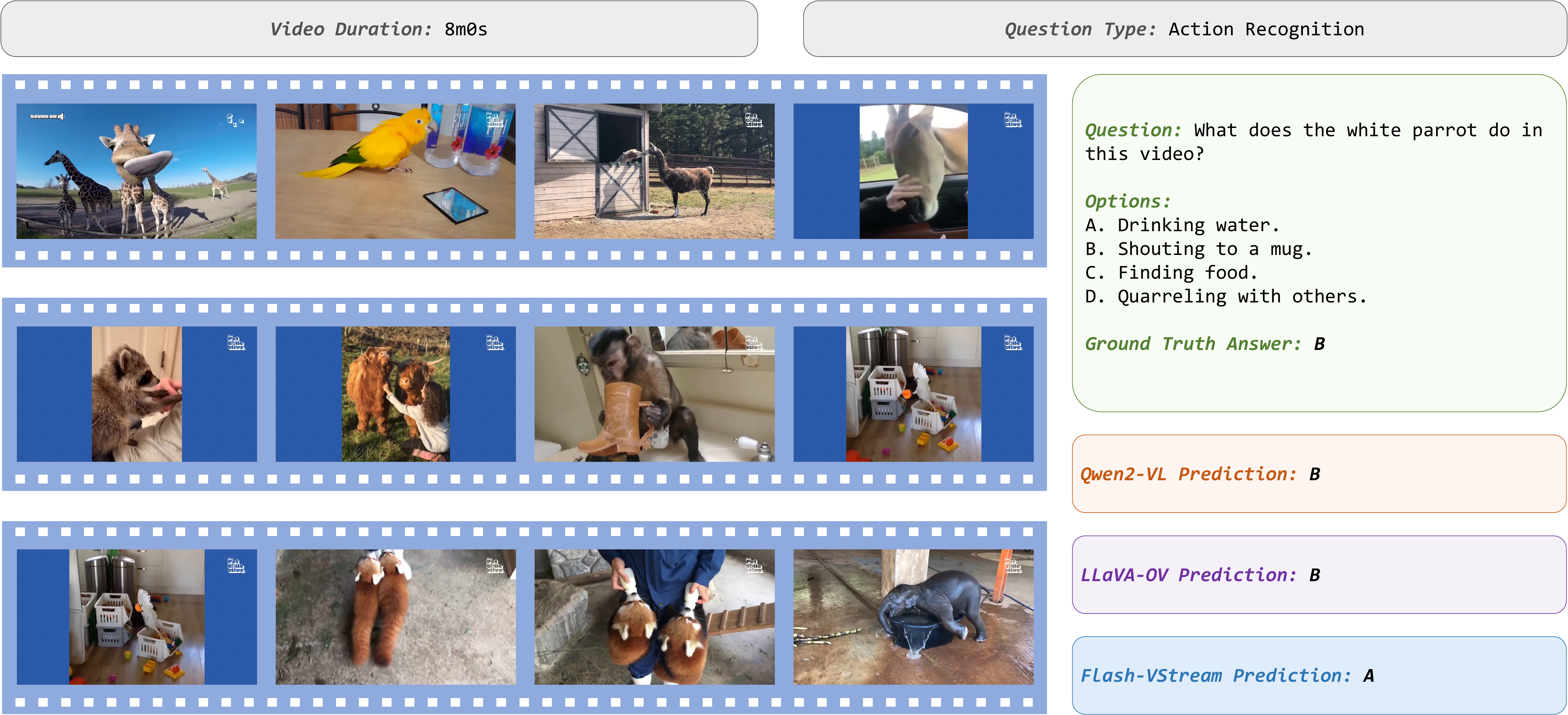}
    \caption{\textbf{Fail Case Analysis.}
    This figure presents a case study involving a video showing various animals and their behaviors. The video frames capture different moments of animal interactions and activities, highlighting the diverse behaviors exhibited by the animals. The study includes a question about the specific action of a white parrot in the video, with multiple-choice options provided. The ground truth answer is indicated, along with the predictions from three different models: Qwen2-VL, LLaVA-OV, and Flash-VStream.
    }
    \label{fig:supp_fail_case_2}
\end{figure*}

\section{Limitations}

\subsection{Fail Case Analysis}
Flash-VStream may produce incorrect predictions in certain scenarios,
such as text-intensive long videos (see~\Cref{fig:supp_fail_case_1}) and long videos with rapid scene changes (see~\Cref{fig:supp_fail_case_2}).
We suggest that these heavily edited video have a different information density distribution compared to native videos, making efficient timing modeling more difficult.

% As shown in~\Cref{fig:supp_fail_case_1}, knowledge-intensive videos require advanced OCR ability and knowledge organization ability, which remain challenging. Another example is videos with highly substantive content, as depicted in~\Cref{fig:supp_fail_case_2}, where rapid scene changes lead to a performance drop of clustering-based methods.

\subsection{GPT-3.5-based Metric for Open-ended VQA}
It is worth noting that some previous works~\cite{moviechat,chatunivi,llamavid} follow Video-ChatGPT~\cite{video-chatgpt} to test models on open-ended VQA 
benchmarks~\cite{yu2019activitynet,xiao2021next}
based on GPT-3.5-based judgment (GPT accuracy and GPT score). 
% Since GPT APIs are proprietary and upgrade over time, this evaluation approach lacks reliability, stability and reproducibility~\cite{li2024seedbench}. 
% Video-ChatGPT~\cite{video-chatgpt} proposed GPT-3.5-based evaluation metrics (GPT accuracy and GPT score) for open-ended VQA benchmarks~\cite{yu2019activitynet,xiao2021next}.
However, we notice that these metrics are highly unstable and prone to bias, so we try to avoid evaluating models on these \textit{LLM-as-a-judge} benchmarks.
Since GPT APIs are proprietary and upgrade over time, this evaluation approach lacks reliability, stability and reproducibility~\cite{li2024seedbench}. 
Furthermore, the evaluation can be disturbed by the hallucination of GPT, leading to a biased evaluation result~\cite{moviechat}.
As presented in~\Cref{tab:supp_gpt35}, there is always a discrepancy between the distribution of GPT accuracy and GPT score. 
Therefore, it is still challenging to benchmark the open-ended VQA ability of MLLMs. 

% Specifically, for answers classified as ``Wrong'', many of them are assigned with a high score like ``4'' or ``5''. This abnormal phenomenon reduces the credibility of this ``$0 \sim 5$ score'' metric in GPT-3.5-based evaluation for open-ended VQA.

\section{Future Work}
Future work could focus on enhancing the models' ability to understand edited videos with intensive text or rapid scene transitions, while maintaining the overall efficiency.
Another interesting direction for future work would be to investigate reliable evaluation methods for open-ended VQA.
Additionally, the techniques developed in this study could be adapted for use in other fields such as robotics and surveillance systems.
We hope that our work will inspire further innovations and improvements in these fields, ultimately leading to more intelligent and versatile systems.
%     {
%         \small
%         \bibliographystyle{ieeenat_fullname}
%         \bibliography{main}
%     }
\end{document}